\documentclass[dvipsnames]{article}

\usepackage[final]{hunyuan}

\usepackage{amsmath}
\usepackage{amssymb}
\usepackage{mathtools}
\usepackage{amsthm}
\usepackage{multirow}
\usepackage{makecell}
\usepackage{subcaption}
\usepackage{wrapfig}
\usepackage{graphicx}

\usepackage{pifont}
\usepackage{fvextra}
\usepackage{xspace}

\usepackage[utf8]{inputenc} %
\usepackage[T1]{fontenc}    %
\usepackage{url}            %
\usepackage{booktabs}       %
\usepackage{amsfonts}       %
\usepackage{nicefrac}       %
\usepackage{microtype}      %
\usepackage{colortbl}
\usepackage{color, soul}
\usepackage[most,breakable]{tcolorbox}
\usepackage{textgreek}
\usepackage{xcolor}

\usepackage[labelfont=bf]{caption}
\usepackage{enumitem}
\usepackage{sectsty}
\usepackage{pgfplots}
\usepackage{fancyhdr}
\usepackage{xcolor} %
\usepackage{caption}
\usepackage{booktabs}
\usepackage{multirow}
\usepackage{graphicx}
\usepackage[table]{xcolor} %

\usepackage{natbib} %
\let\cite\citep    %

\renewcommand{\headrulewidth}{1pt}
\makeatletter
\def\headrule{{\if@fancyplain\let\headrulewidth\plainheadrulewidth\fi
\hrule\@height\headrulewidth\@width\textwidth \vskip-\headrulewidth}}
\makeatother

\definecolor{HYDarkBlue}{HTML}{2155EA} 
\definecolor{HYLightBlue}{HTML}{A8DFF6}
\usepackage[colorlinks, linkcolor=HYDarkBlue, anchorcolor=HYDarkBlue, citecolor=HYDarkBlue, urlcolor=HYDarkBlue]{hyperref}

\usepackage{cleveref}
\usepackage{threeparttable}

\sectionfont{\color{HYDarkBlue}\fontfamily{zi4}\selectfont}
\subsectionfont{\color{HYDarkBlue}\fontfamily{zi4}\selectfont}
\subsubsectionfont{\color{HYDarkBlue}\fontfamily{zi4}\selectfont}

\newtcolorbox{mytheorem}{
  colback=gray!5,       %
  colframe=gray!80,     %
  boxrule=0.5pt,        %
  arc=4pt,              %
  left=4pt,             %
  right=4pt,            %
  top=4pt,              %
  bottom=4pt,           %
}

\newcommand{\fancyheadname}{\textit{\textbf{\modelname{}}}}

\title{AngelSlim: A more accessible, comprehensive, and efficient toolkit for large model compression}

\author{%
\\
\textbf{\large{Hunyuan AI Infra Team}}
\vspace{0em}
}

\def\huggingface{\raisebox{-1.5pt}{\includegraphics[height=1.05em]{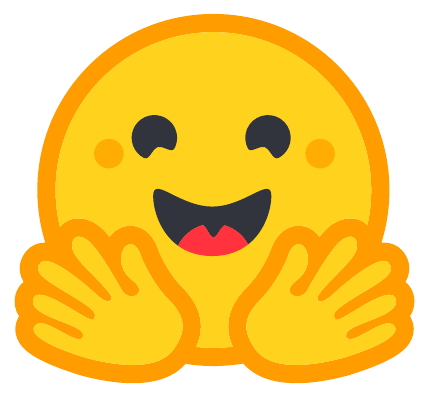}}}
\def\github{\raisebox{-1.5pt}{\includegraphics[height=1.05em]{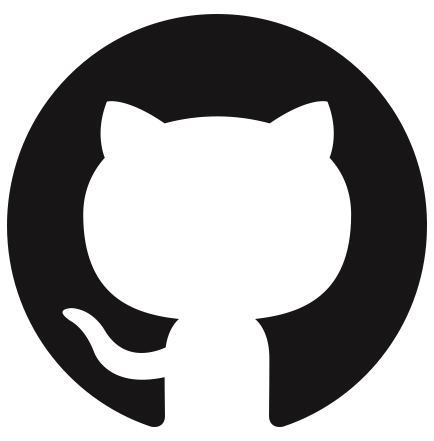}}}
\def\email{\raisebox{-1.5pt}{\includegraphics[height=1.05em]{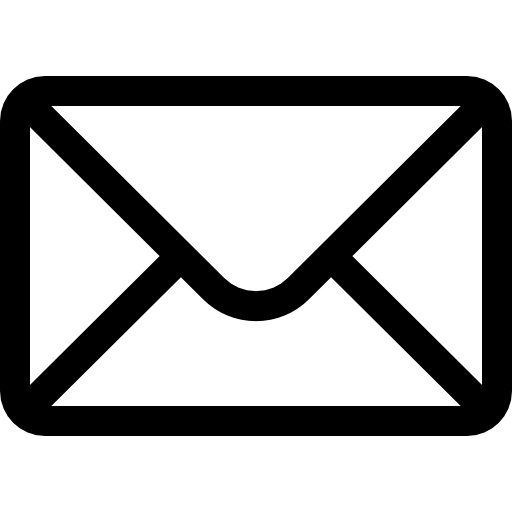}}}
\newcommand{\modelname}[0]{AngelSlim}

\usepackage{geometry}
\usepackage{algorithm}
\usepackage{algpseudocode}
\usepackage{setspace}

\makeatletter
\renewcommand{\ALG@beginalgorithmic}{\small}
\makeatother



\begin{document}

\maketitle
\thispagestyle{fancy} %

\begin{abstract}
This technical report introduces \textbf{\modelname{}}, a comprehensive and versatile toolkit for large model compression developed by the Tencent Hunyuan team. By consolidating cutting-edge algorithms, including quantization, speculative decoding, token pruning, and distillation. \modelname{} provides a unified pipeline that streamlines the transition from model compression to industrial-scale deployment. To facilitate efficient acceleration, we integrate state-of-the-art FP8 and INT8 Post-Training Quantization (PTQ) algorithms alongside pioneering research in ultra-low-bit regimes, featuring \textbf{HY-1.8B-int2} as the first industrially viable 2-bit large model. 
Beyond quantization, we propose a training-aligned speculative decoding framework compatible with multimodal architectures and modern inference engines, achieving $1.8\times$ to $2.0\times$ throughput gains without compromising output correctness. Furthermore, we develop a training-free sparse attention framework that reduces Time-to-First-Token (TTFT) in long-context scenarios by decoupling sparse kernels from model architectures through a hybrid of static patterns and dynamic token selection. For multimodal models, \modelname{} incorporates specialized pruning strategies, namely IDPruner for optimizing vision tokens via Maximal Marginal Relevance and Samp for adaptive audio token merging and pruning. By integrating these compression strategies from low-level implementations, \modelname{} enables algorithm-focused research and tool-assisted deployment.

\end{abstract}

\begin{figure}[h]
    \centering
\includegraphics[width=1.0\linewidth]{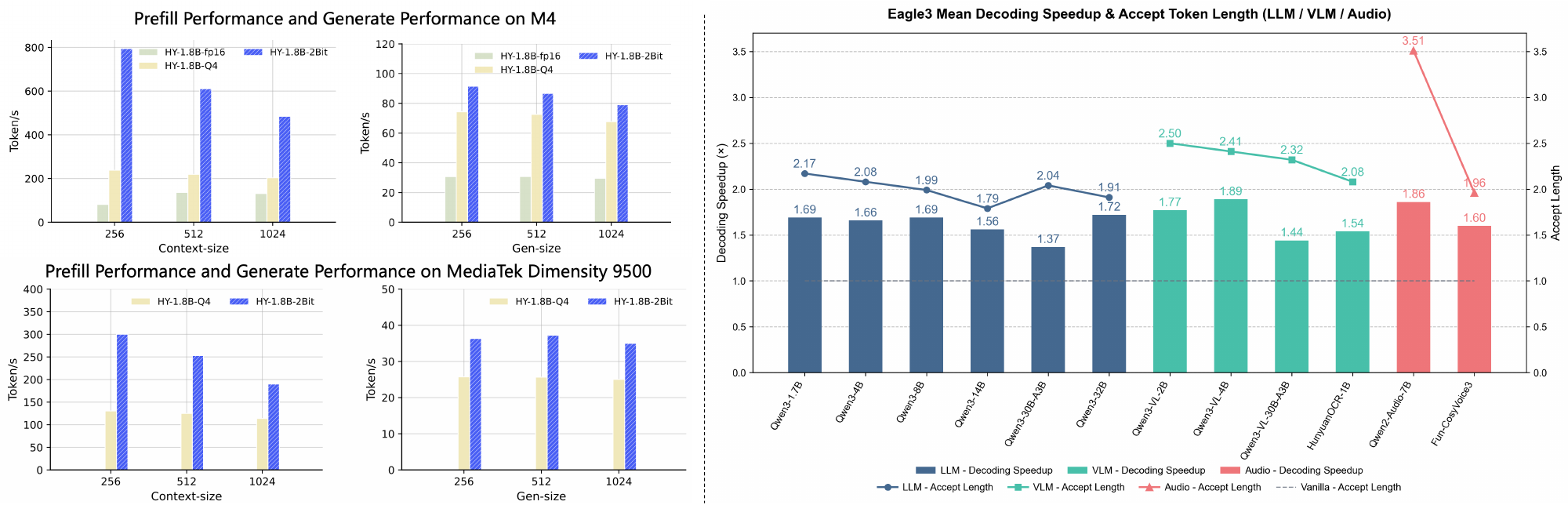}
    \caption{\textit{(Left)} the edge effciency of HY-1.8B-2bit. \textit{(Right)} the speedup of speculative sampling.}
    \label{fig:efficiency-eval}
\end{figure}

\newpage
{
  \hypersetup{linkcolor=RoyalBlue, linktoc=page}
  \tableofcontents
}

\newpage

\section{Introduction} 
The landscape of Artificial Intelligence has been fundamentally redefined by the emergence of large-scale foundation models~\cite{achiam2023gpt,bai2023qwen, wu2025qwen,cao2025hunyuanimage,liu2024deepseek}. Large Language Models (LLMs)~\cite{achiam2023gpt,liu2024deepseek,bai2023qwen} and Multimodal Large Models (MLMs)~\cite{wu2025qwen,cao2025hunyuanimage} have demonstrated unprecedented capabilities in complex reasoning, zero-shot generalization, and cross-modal synthesis. However, these advancements are inextricably linked to a significant computational paradox known as the ``Inference Wall.'' The deployment of these models is severely hindered by the quadratic scaling of the self-attention mechanism~\cite{vaswani2017attention} and the massive memory bandwidth required to fetch billions of parameters. As the demand for longer context windows and real-time multimodal interaction grows, the gap between the hardware requirements of state-of-the-art (SOTA) models and the constraints of commodity GPUs creates a critical bottleneck for the democratization and practical application of AI.

To mitigate these overheads, the research community has proposed several compression paradigms, including quantization~\cite{frantar2022gptq,lin2024awq}, sparsification~\cite{yuan2025native}, and speculative decoding~\cite{leviathan2023fast}. However, existing research typically treats these optimizations as isolated silos. This fragmentation prevents the realization of an end-to-end compression pipeline, as individual methods often fail to account for the cross-technique interference that arises when they are applied concurrently. Consequently, there remains a significant gap between theoretical compression and practical, unified deployment.

In this paper, we present \textbf{AngleSlim}, a unified compression and acceleration framework designed to bridge the gap between disparate optimization techniques. Unlike prior works~\cite{frantar2022gptq,lin2024awq,yuan2025native}\nocite{huang2025quaff} that prioritize a single axis of efficiency, AngleSlim treats quantization, speculative decoding, sparse attention, and token pruning as a cohesive, multi-objective manifold. By synchronizing weight-level precision with dynamic structural sparsity and algorithmic look-ahead, AngleSlim enables large-scale LLMs and MLMs to operate at a fraction of their original computational cost without sacrificing reasoning integrity or multimodal alignment. This holistic integration allows for a hardware-aware optimization process that pushes the boundaries of the inference efficiency frontier.

The first pillar of AngleSlim is a multi-tier quantization suite that addresses the precision-efficiency gap through both Post-Training Quantization (PTQ) and Quantization-Aware Training (QAT)~\cite{liu2025paretoq}. In the QAT regime, we first incorporate Stretched Elastic Quantization~\cite{liu2025paretoq}, an aggressive ultra-low-bit strategy that enables the compression of weights into a 2-bit representation. This allows for the development of highly efficient on-device models, such as our HY-1.8B-2Bit model. Furthermore, we push the boundaries of extreme compression by introducing two novel ternary quantization strategy: \textbf{1.58-Bit Tequila}~\cite{huang2025tequila}, a trapping-free ternary quantization method, and \textbf{1.25-Bit Sherry}~\cite{huang2026sherry}, which achieves hardware-efficient 3:4 sparse ternary quantization. In the PTQ regime, AngelSlim introduces Leptokurtic Quant (\textbf{LeptoQuant}) with  Dynamic Outlier Isolation Scale. By observing that weight distributions often follow a Laplacian-like peak with significant outliers, LeptoQuant mitigates the performance degradation typically seen in standard FP8-E4M3 quantization. Instead of allowing traditional smoothing to erode precision in densely populated numerical ranges, LeptoQuant searches for optimal scaling values to isolate outliers and preserve the representational fidelity of the original distribution, ensuring that even the most compressed weights retain high task performance.

Beyond quantization, AngelSlim further targets inference-time efficiency through a speculative sampling–oriented training framework that is explicitly designed to align draft models with their corresponding target models. This framework enables efficient and stable training of Eagle3-style speculative decoders, ensuring high token acceptance rates without sacrificing output correctness. The training pipeline is multi-modal, scalable, and production-oriented, supporting language, vision–language, OCR, and speech models, and is directly deployable in modern inference engines such as vLLM. Across a wide range of benchmarks, AngelSlim-trained draft models consistently deliver substantial end-to-end acceleration: Eagle3~\cite{li2025eagle} speculative decoding achieves an average throughput improvement of 1.8–2.0×, while maintaining correctness, with the average number of accepted speculative tokens (AL) ranging from 1.7 to 3.5, depending on model size and modality. These results demonstrate that AngelSlim enables speculative decoding not merely as an inference-time heuristic, but as a training-aligned system capability. Complementary algorithmic advances, such as SpecExit ~\cite{yang2025specexit}, further illustrate the broader design space for dynamic early-exit and adaptive computation, highlighting potential directions for integrating speculative decoding with fine-grained reasoning control in large models.

Moreover, AngelSlim features a training-free sparse attention framework designed to mitigate the computational intensity of long-context inference.
By exploiting the inherent sparsity within attention mechanisms, this framework significantly accelerates the prefill process during the KV Cache Generation stage.
The system provides a dual-optimization strategy encompassing both static and dynamic regimes.
In the static regime, structural heuristics such as A-shape, Tri-shape, and Dilated patterns are employed to maintain a broad receptive field with fixed sparsity masks.
In the dynamic regime, AngelSlim integrates advanced algorithms—including MInference~\citep{minference}, XAttention~\citep{xattention}, FlexPrefill~\citep{flexprefill}, and Stem—to selectively compute salient tokens in real-time.
By implementing a strict decoupling between sparse kernels and model architectures through a metadata-driven system, the framework offers seamless acceleration for mainstream models like Hunyuan-Dense-Model~\citep{hunyuandense}, Llama-3.1~\citep{llama31}, and Qwen3~\citep{qwen3}.
This approach effectively slashes the Time-to-First-Token (TTFT) while maintaining high accuracy across long-text benchmarks such as LongBench~\citep{longbench} and RULER~\citep{ruler}.

Specifically for Multi-modal Large Language Models (MLLMs), visual and audio inputs generate a vast number of tokens, many of which exhibit spatial and temporal redundancy.
To address this substantial computational burden, AngelSlim offers token reduction methods for visual and audio modalities, respectively, alongside a unified system framework that enables efficient multimodal acceleration.
For the visual modality, we propose IDPruner, which systematically harmonizes token importance and semantic diversity by adapting the Maximal Marginal Relevance (MMR) principle without requiring attention maps.
For audio inputs, we introduce Samp, a similarity-attention synergistically driven framework that employs a two-stage merging-pruning pipeline and adaptively calibrates the merging-pruning ratio for each sample.
Both methods achieve state-of-the-art performance in their respective modalities and are integrated within a unified system architecture.
Besides, by decoupling pruning strategies from model-specific implementations, AngelSlim enables researchers to concentrate on algorithm design without concerning themselves with low-level implementation details.
The framework facilitates straightforward deployment with automated benchmarking tools for measuring both task accuracy and inference latency.

In summary, \modelname{} establishes a new SOTA for comprehensive model compression across diverse tasks and benchmarks. Our empirical results demonstrate that the proposed ternary quantization and novel 2-bit QAT framework achieve accuracy parity with INT4 precision while delivering a substantial $4\times$ speedup on edge devices. Furthermore, our high-efficiency 4-8 bit PTQ suite ensures near-lossless compression with a low-memory footprint, while the Eagle3 framework provides a seamless, training-aligned pipeline for multimodal speculative decoding. By integrating training-free sparse attention libraries and unified token pruning frameworks for MLLMs and audio tasks, \modelname{} delivers a production-ready infrastructure that maximizes inference throughput without compromising reasoning integrity. As an accessible and exhaustive toolkit, \modelname{} bridges the gap between algorithmic innovation and industrial-scale deployment, with a commitment to continuous functional expansion for the research community.

\section{AngelSlim Quantization Framework}
In this section, we will introduce the 2-bit quantization algorithm of the HY-1.8B-2Bit model and the ternary quantization, covering its data, training strategy, and quantization scheme. We will also introduce AngelSlim's PTQ framework, which supports a variety of 4–8 bit quantization algorithms and provides a one‑click interface for developer convenience. Additionally, we will present LeptoQuant, an FP8 quantization algorithm designed for industrial deployment.

\subsection{HY-1.8B-2Bit : An Efficient On-Device LLM}

HY-1.8B-2Bit  is a high-efficiency 2-bit LLM built for on-device deployment. It is derived from the Tencent Hunyuan-1.8B-Instruct model~\citep{hunyuan1.8b} via an optimized Quantization-Aware Training (QAT) framework~\citep{liu2025paretoq}, while preserving a lightweight dense architecture. To overcome the stringent resource constraints of edge hardware, HY-1.8B-2Bit  utilizes Stretched Elastic Quantization (SEQ)~\cite{liu2025paretoq}, an aggressive ultra-low-bit strategy that compresses weights to a 2-bit representation. This results in a bit-equivalent model size of merely 0.3B, representing a $6\times$ compression compared to the original Hunyuan-1.8B model, which significantly alleviates inference latency and computational overhead. Despite the radical reduction in precision, HY-1.8B-2Bit  retains the sophisticated Dual Chain-of-Thought (Dual-CoT) reasoning capability, achieving a significant average accuracy gain of 17\% across multiple benchmarks compared to models with equivalent model size.

\subsubsection{Model Strengths}

Through the application of optimized quantization-aware training, HY-1.8B-2Bit  achieves a performance profile that is remarkably competitive with the PTQ-INT4 variant, HY-1.8B-Instruct-Int4~\cite{hunyuanint4}. Comprehensive evaluations across diverse domains—including mathematics, humanities, and programming—reveal that HY-1.8B-2Bit  incurs a marginal average performance degradation of only 4\% relative to the full-precision Hunyuan-1.8B-Instruct~\cite{hunyuan1.8b} baseline. Given the $6\times$ compression ratio achieved, this minimal loss highlights the model's exceptional information retention and its capacity to maintain structural robustness even under radical bit-width reduction.

\paragraph{Superior Model Capability} 
Through the application of QAT, HY-1.8B-2Bit  achieves a performance profile that is remarkably competitive with the PTQ-INT4 variant, Hunyuan-1.8B-Instruct-INT4~\cite{hunyuanint4}.  Comprehensive evaluations across diverse domains, including mathematics, humanities, and programming, reveal that HY-1.8B-2Bit  incurs a marginal average performance degradation of only 4\% relative to the full-precision Hunyuan-1.8B-Instruct~\cite{hunyuan1.8b} baseline. Given the $6\times$ compression, this minimal loss highlights the model's exceptional information retention and its capacity to maintain structural robustness even under radical bit-width reduction.

\paragraph{Efficiency Gains with Lightweight Scale}
HY-1.8B-2Bit  redefines the efficiency frontier for small-scale language models. When benchmarked against dense models of equivalent bit-equivalent size, such as HY-0.5B-Instruct~\cite{hunyuan0.5b}, HY-1.8B-2Bit  maintains a substantial lead, outperforming its counterpart by an average of 16\% across standard benchmarks. This performance gap suggests that knowledge from a larger model into an ultra-low-bit representation is more effective than training a smaller dense model from scratch. Consequently, HY-1.8B-2Bit  serves as a scalable, high-efficiency alternative for edge computing, delivering sophisticated reasoning capabilities within a hardware-friendly footprint.

\paragraph{Comprehensive Reasoning Proficiency}
A distinguishing feature of HY-1.8B-2Bit  is its inheritance of the "full-thinking" architecture from the HY-A13B~\cite{hunyuana13b}, making it the most compact model in the industry to support complex reasoning pathways. By implementing a Dual Chain-of-Thought (Dual-CoT) strategy, HY-1.8B-2Bit  allows for a dynamic trade-off between execution speed and cognitive depth. Users can invoke a concise "short-CoT" for straightforward inquiries to minimize latency, or a detailed "long-CoT" for tasks requiring rigorous logical multi-step synthesis. This dual-mode flexibility ensures that HY-1.8B-2Bit is uniquely suited for real-time, resource-constrained deployments where both high-fidelity logic and rapid response are non-negotiable requirements.

\subsubsection{2-bit QAT Workflow}
We present a high-efficiency QAT methodology specifically designed for HY-1.8B-2Bit. Our framework demonstrates the empirical feasibility of compressing high-precision LLMs into ultra-low bit-widths without incurring the "accuracy collapse" typically associated with extreme quantization. By leveraging low-bit QAT, we demonstrate that a model's capacity-to-size ratio can be elevated beyond the theoretical limits of traditional full-precision models of equivalent parameter counts, enabling the deployment of sophisticated intelligence in strictly resource-constrained environments.

\paragraph{Data Curative Pipeline.} To ensure high-fidelity knowledge retention, we adapt the hierarchical data selection pipeline established in HY-A13B~\cite{hunyuana13b}. This pipeline consists of three core stages: (1) Preprocessing: involving global deduplication, heuristic-based quality filtering, and denoising; (2) Model-Based Extraction: utilizing specialized classifiers to distill coherent text from raw corpora; and (3) Refinement: which applies semantic-level deduplication and rigorous quality assessments.

We hypothesize that a compact, high-signal subset is superior to a massive, noisy corpus in the QAT process. Our preliminary experiments and previous works~\cite{li2025quantization} indicate that performance decay in ultra-low-bit regimes is disproportionately concentrated in logical reasoning and long-context comprehension. To mitigate this, we curated a specialized hybrid dataset by increasing the proportion of scientific and mathematical tokens and integrating targeted long-form sequences. This resulted in \textbf{an optimized set of 89B tokens}, providing a high-density signal for recovering the model's cognitive capabilities during quantization.

\paragraph{Quantization Strategy}
Addressing the inherent instability of 2-bit precision, HY-1.8B-2Bit employs SEQ~\cite{liu2025paretoq} to stabilize training. Traditional asymmetric INT2 mappings that include a zero value (e.g., $\{-2, -1, 0, 1\}$) often suffer from restricted dynamic range. In contrast, SEQ adopts a symmetric mapping scheme: $\{-1.5, -0.5, 0.5, 1.5\}$. By shifting the quantization centroid and eliminating the zero-point, SEQ optimizes the dynamic range coverage and resolves the "limited energy level" bottleneck. Coupled with an adaptive micro-tuning of the scaling factor for quantization intervals, this strategy significantly mitigates information dissipation, capturing high-dimensional features even within a 2-bit constraint.

\paragraph{QAT Configuration and Optimization}
In the 2-bit regime, QAT transitions from a simple "error compensation" task to a complex "distribution reconstruction" process, where weight manifolds undergo significant shifts to align with low-precision representations.

\begin{itemize}
    \item \textbf{Initialization and Convergence:} Unlike frameworks such as BitNet~\cite{ma2025bitnet}, we initialize HY-1.8B-2Bit with instruction-tuned weights rather than raw pre-trained weights. This strategic initialization provides a superior starting point for the manifold reconstruction, accelerating convergence and drastically reducing the token budget required for recovery.
    \item \textbf{Hyperparameter "Wind Tunnel" Testing:} To navigate the sensitivity of 2-bit landscapes, we conducted extensive small-scale "wind tunnel" experiments using a 10B token subset. This allowed us to rapidly isolate optimal hyperparameters, bypassing the prohibitive computational cost of full-scale grid searches.
    \item \textbf{Data Efficiency:} We analyzed the performance returns of varying data volumes (30\%, 50\%, and 70\% of the full SFT corpus). Our findings indicate that 50\% of the SFT data provides an optimal equilibrium between accuracy recovery and computational overhead.
\end{itemize}

Notably, HY-1.8B-2Bit requires only 10\% of the token consumption compared to models like BitNet-2B~\cite{ma2025bitnet}. This efficiency demonstrates that high-performance, ultra-low-bit models do not necessitate training from scratch, providing a scalable and cost-effective blueprint for the industrial-scale production of quantized LLMs.

\subsubsection{Performance and Deployment Efficiency of HY-1.8B-2Bit}

We evaluate HY-1.8B-2Bit across diverse standard benchmarks to assess its general knowledge (CMMLU~\cite{li2024cmmlu}, C-Eval~\cite{huang2023c}, GPQA Diamond Pass@3~\cite{rein2024gpqa}), logical reasoning (ARC~\cite{clark2018think}, BBH~\cite{suzgun2023challenging}), and specialized technical proficiency (GSM8K~\cite{cobbe2021gsm8k}, HumanEval Pass@3~\cite{chen2021evaluating}, LiveCodeBench~\cite{jain2024livecodebench}). Performance is measured against the full-precision HY-1.8B model~\cite{hunyuan1.8b}, an INT4 baseline~\cite{hunyuanint4} quantized via GPTQ~\cite{frantar2022gptq}, and a scale-equivalent dense model, HY-0.5B~\cite{hunyuan0.5b}. 

\begin{table}[t]
\centering
\resizebox{\textwidth}{!}{%
\begin{tabular}{lcccccccccc}
\toprule
\textbf{Model} & \textbf{CMMLU} & \textbf{C-Eval} & \textbf{ARC} & \textbf{BBH} & \textbf{GSM8K} & \textbf{HumanEval} & \textbf{LCB} & \textbf{GPQA} & \textbf{Average} & \textbf{Distance} \\ \midrule
HY-1.8B-FP16 & 55.07\% & 54.27\% & 70.50\% & 79.08\% & 84.08\% & 94.51\% & 31.50\% & 68.18\% & 67.15\% & 0.00\% \\
HY-0.5B-FP16 & 37.08\% & 35.98\% & 49.89\% & 58.10\% & 55.04\% & 67.07\% & 12.11\% & 46.97\% & 45.28\% & -21.87\% \\
HY-1.8B-INT4 & 50.80\% & 48.67\% & 68.83\% & 74.80\% & 78.70\% & 89.02\% & 30.08\% & 65.56\% & 63.31\% & -3.84\% \\
\rowcolor{HYLightBlue} \textbf{HY-1.8B-2Bit} & 49.32\% & 47.60\% & 64.45\% & 75.54\% & 77.33\% & 93.29\% & 32.73\% & 65.15\% & 63.18\% & -3.97\% \\ \bottomrule
\end{tabular}%
}
\vspace{0.5em}
\caption{Benchmark performance comparison across diverse domains. All models are instruction-tuned. The "Distance" column denotes the  accuracy gap relative to the 1.8B FP16 baseline.}
\label{tab:performance}
\end{table}

\begin{figure}[t]
    \centering
    \includegraphics[width=0.9\linewidth]{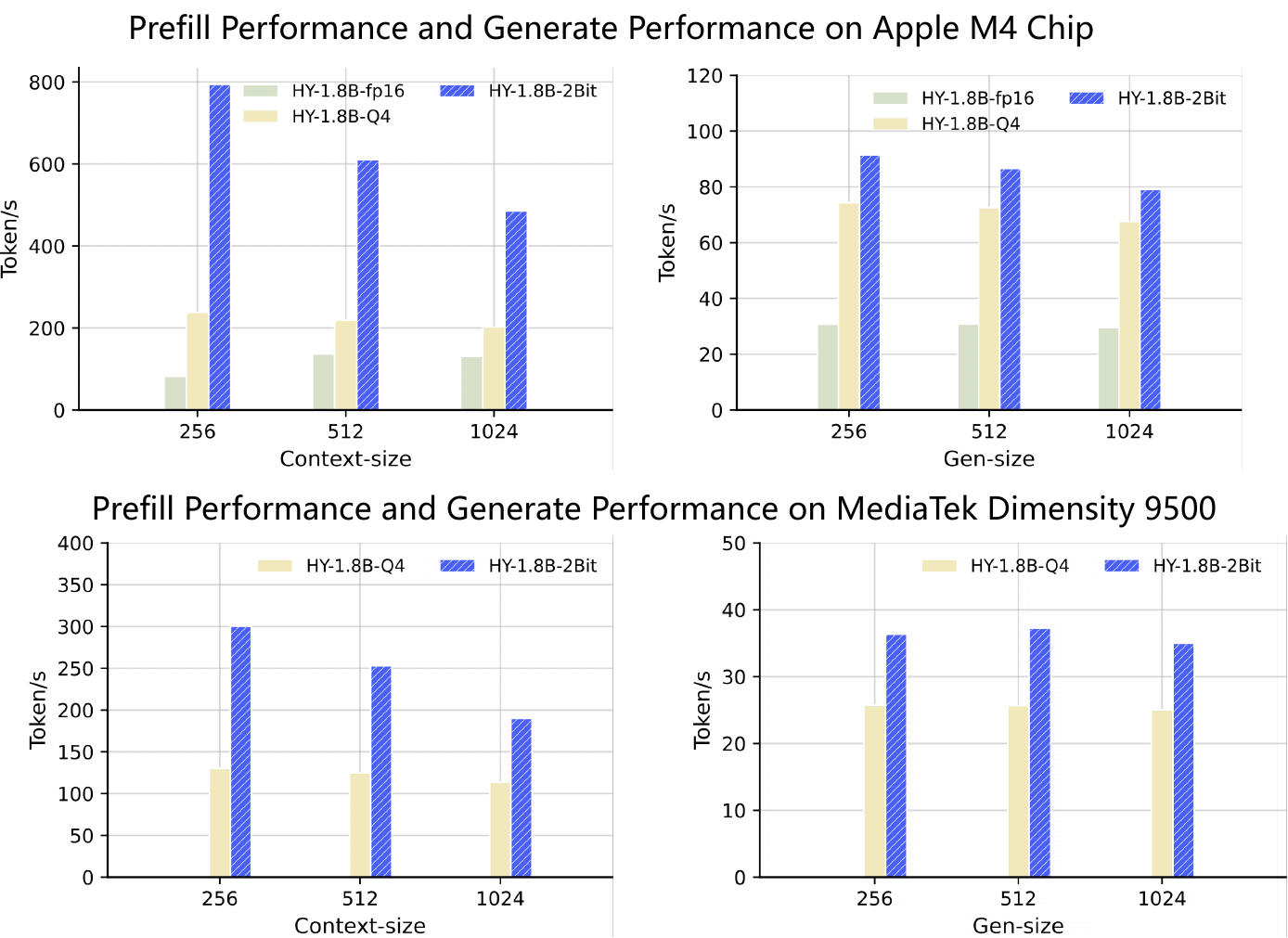}
    \caption{Comparison of inference latency (TTFT) and generation throughput on edge devices. HY-1.8B-2Bit consistently outperforms both 4-bit and FP16 baselines across varying sequence lengths.}
    \label{fig:eff-nano}
\end{figure}

\paragraph{Performance of HY-1.8B-2Bit:}
Table~\ref{tab:performance} summarizes the comparative performance of HY-1.8B-2Bit. The empirical results reveal that HY-1.8B-2Bit maintains high-tier performance despite the extreme reduction in bit-width, incurring a marginal average degradation of only 3.97\% compared to its full-precision 1.8B teacher. Remarkably, HY-1.8B-2Bit performs nearly on par with the INT4 variant,with a negligible accuracy gap of only 0.13\%, while utilizing only half the weight precision. In reasoning-intensive benchmarks such as BBH and LiveCodeBench, HY-1.8B-2Bit actually outperforms the 4-bit baseline. This suggests that our QAT framework effectively reconstructs the high-dimensional weight distributions necessary to preserve the complex logical structures of the base model, even under 2-bit constraints. When compared to the dense HY-0.5B model, which occupies a comparable model size, the superiority of the 2-bit QAT approach becomes evident.  While the 0.5B dense model suffers a catastrophic 21.87\% drop in average accuracy, HY-1.8B-2Bit remains robust, outperforming the smaller dense counterpart by 22.29\% in GSM8K and 20.62\% in LiveCodeBench. This significant gap validates our strategy of compressing a larger, high-capacity model into an ultra-low-bit format rather than training a smaller parameter model from scratch. This makes HY-1.8B-2Bit a competitive alternative for edge-based intelligence where memory and power budgets are strictly capped.

\paragraph{Efficiency of HY-1.8B-2Bit on Edge Hardware:} We evaluate the practical deployment efficiency of HY-1.8B-2Bit on both Apple M4 and MediaTek Dimensity 9500 Chip. To standardize the benchmarking environment, we utilized a fixed configuration of 2 threads to measure Time-to-First-Token (TTFT) and generation throughput across varying context window sizes. For comparative analysis, we benchmarked the FP16, Q4\_K\_M (4-bit), and our HY-1.8B-2Bit GGUF formats. As illustrated in Figure~\ref{fig:eff-nano}, HY-1.8B-2Bit demonstrates substantial gains on the Apple M4 chip, achieving a $3\times$ to $8\times$ acceleration in TTFT for input sequences ranging from 256 to 1024 tokens compared to the BF16 baseline. Regarding generation speed, HY-1.8B-2Bit maintains a consistent $>2\times$ speedup over BF16 across standard operating windows. On the MediaTek Dimensity 9500 Chip, the HY-1.8B-2Bit achieves a $\sim2\times$ speedup compared to 4-bit model in prefill performance, while maintaining a $\sim1.5\times$ in generative throughput. These results underscore HY-1.8B-2Bit's capacity for high-performance, real-time inference in resource-constrained environments.

\begin{figure}[t]
    \centering
    \includegraphics[width=0.9\linewidth]{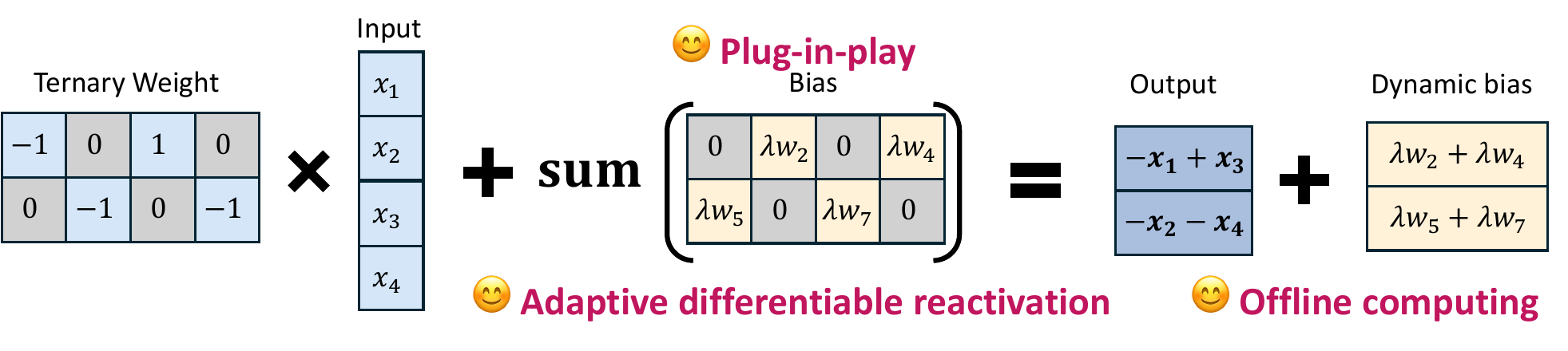}
    \caption{Tequila reactivates dead weights as adaptive dynamic biases via a differentiable function, achieving significant accuracy improvements with nearly zero inference overhead.}
    \label{fig:tequila}
\end{figure}

\subsection{Ternary Quantization}
The AngelSlim framework incorporates ternary quantization as a core strategy for extreme weight compression, constraining model weights to the discrete set $\{-1, 0, +1\}$. This paradigm effectively replaces computationally intensive floating-point multiplications with hardware-efficient additions via a lookup table-based engine like BitNet.cpp~\cite{wei2025t} and T-MAC~\cite{wang2025bitnetcpp}. To overcome the traditional trade-offs between model accuracy and hardware alignment, AngelSlim introduces two specialized methodologies: {\color{HYDarkBlue}\textbf{Tequila}}~\cite{huang2025tequila} and {\color{HYDarkBlue}\textbf{Sherry}}~\cite{huang2026sherry}.

\subsubsection{1.58-Bit Tequila: Trapping-Free Ternary Quantization}
Tequila addresses the \textit{deadzone trapping} phenomenon inherent in standard ternary QAT. In conventional ternary quantization schemes~\cite{ma2025bitnet}, the quantization function for a weight vector $W = (w_1, \dots, w_n)$ is typically defined as:
\begin{equation}
   Q(W) = \hat{W}\alpha, \quad \hat{w}_i = \begin{cases}
    +1, & \text{if } w_i \geq \Delta ; \\ 
    0, & \text{if } |w_i| < \Delta; \\ 
   -1, & \text{if } w_i \le -\Delta , 
  \end{cases} 
  \label{eq:tquant}
\end{equation}
where $\hat{W}$ represents the ternary weights, $\alpha$ is a scaling factor, and $\Delta$ is the threshold. Due to the coarse nature of the Straight-Through Estimator (STE), weights within the "deadzone" $[-\Delta, \Delta]$ often receive uninformative gradients, leading to representational stagnation.

As illustrated in Figure~\ref{fig:tequila}, Tequila reactivates these trapped weights by repurposing them as  biases during training:
 \begin{equation}
Y = XQ(W) + C(W) = X\hat{W}\alpha + \sum_{i \in D}\lambda w_i,
\label{eq:tequila}
\end{equation}
where $D = \{i \mid |w_i| < \Delta\}$ denotes the set of indices in the deadzone. The bias term $C(W)$ acts as a residual connection, yielding an informative gradient signal for the dead weights:
\begin{equation}
    \frac{\partial L}{\partial w_i} = x_i\frac{\partial L}{\partial Y} + \lambda\frac{\partial L}{\partial Y}, \quad \forall i \in D.
    \label{eq:mixgrad}
\end{equation}
This formulation enables weights to escape the deadzone stably by providing a continuous signal in the forward pass and a direct gradient path in the backward pass. Crucially, these biases are merged into static network parameters offline post-training, ensuring nearly zero inference overhead while maintaining the theoretical ternary quantization.

\begin{figure}[h]
    \centering
    \includegraphics[width=0.9\linewidth]{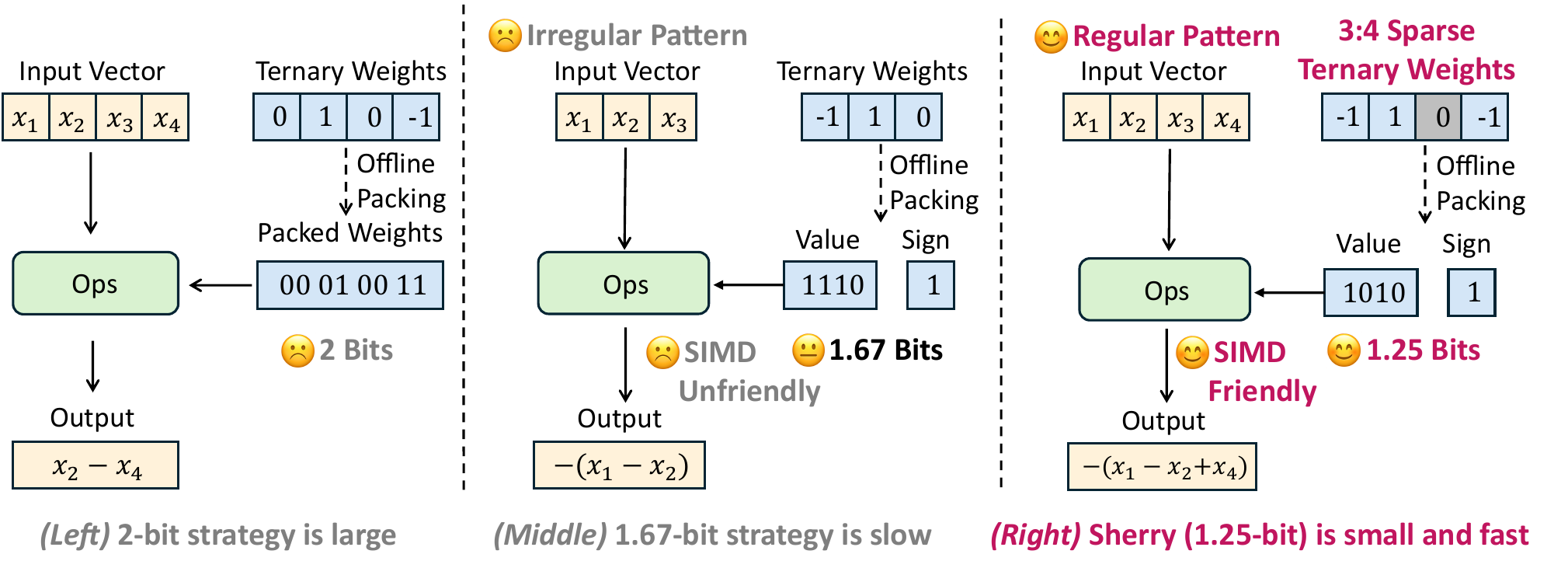}
    \caption{\textit{(Left)} \textbf{2-bit strategy} packs each weight into 2 bits to maintain alignment, resulting in large bit wastage.
    \textit{(Middle)} \textbf{ 1.67-bit strategy} packs 3 weights into 5 bits, introducing SIMD-unfriendly 3-way patterns, leading to slow speed. 
    \textit{(Right)} \textbf{Our Sherry} enforces a 3:4 sparsity and packs 4 weights into 5 bits,  introducing SIMD-friendly 4-way patterns, achieving a small 1.25-bit width and faster inference speed.}
    \label{fig:sherry}
\end{figure}

\subsubsection{1.25-Bit Sherry: Hardware-Efficient 3:4 Sparsification}
Sherry resolves the architectural friction between non-standard ternary bit-widths and commodity hardware by introducing \textbf{3:4 fine-grained structured sparsity}. While standard ternary methods often force a choice between bit wastage (2-bit packing) or reduced inference throughput (1.67-bit irregular packing), Sherry restores power-of-two alignment,as shown in Figure~\ref{fig:sherry}.

\paragraph{3:4 Sparse Packing}
Sherry enforces a structural constraint where exactly three non-zero elements ($\pm 1$) are permitted within every contiguous block of four weights. This configuration allows for an optimal packing strategy where each block is stored in a compact 5-bit representation, as $N_{perm} = \binom{4}{3} \times 2^3 = 32$, which perfectly saturates a 5-bit index. This regularized mapping achieves a 1.25-bit width while maintaining native compatibility with Single Instruction Multiple Data (SIMD) vector lanes. 
\begin{wrapfigure}{r}{0.45\textwidth} 
    \centering
    \includegraphics[width=0.9\linewidth]{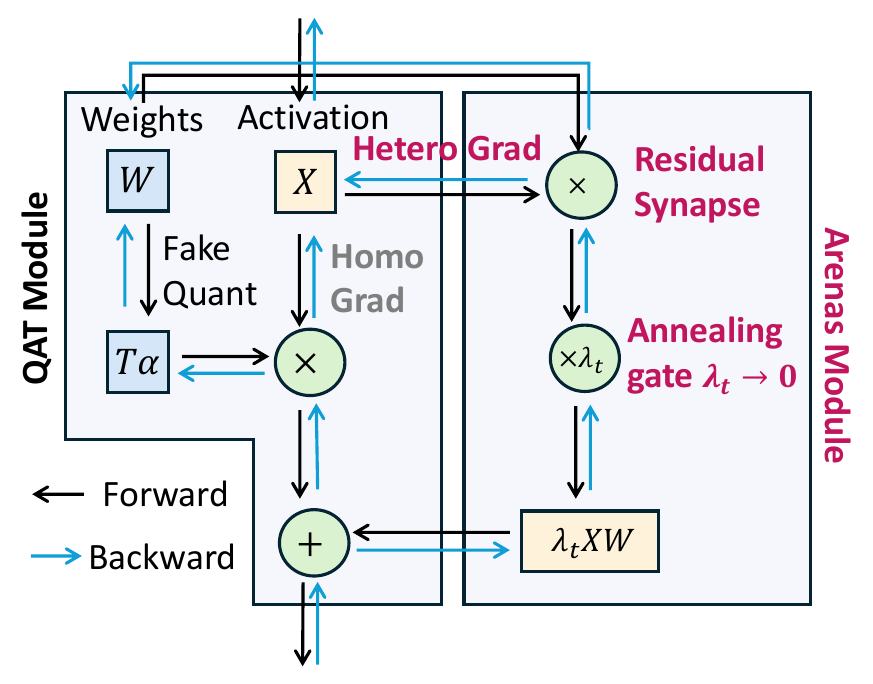}
    \caption{The overview of the Arenas module with QAT. The Arenas module injects the heterogeneous gradients through a residual synapse with an annealing gate.}
    \label{fig:Arenas}
    \vspace{-1em}
\end{wrapfigure}

\paragraph{Arenas: Annealing Residual Synapse} 
Directly imposing 3:4 sparsity on ternary weights frequently leads to \textit{weight trapping}, where gradients become homogenized and weights accumulate in localized regions, causing representational collapse. To counteract this, Sherry introduces \textbf{Arenas}, a module that injects heterogeneous gradients during the training phase:
\begin{equation}
    Y = XQ(W) + \lambda_tXW,
\end{equation}
where $\lambda_t$ is a scheduling coefficient that anneals to zero by the conclusion of training. By re-introducing the continuous latent weight $W$ into the forward pass, Arenas prevents the gradients $\frac{\partial L}{\partial X}$ from collapsing into a low-rank state. This encourages diversity in weight updates and ensures the final sparse model retains high expressive capacity without any additional inference-time cost.

\subsubsection{Performance and Deployment Efficiency}
We evaluate Tequila and Sherry on the LLaMA-3.2 1B and 3B models across multiple standard benchmarks to demonstrate their superior accuracy-efficiency trade-offs. The evaluation covers five zero-shot tasks: PIQA~\citep{bisk2020piqa}, ARC-Easy (ARC-e), ARC-Challenge (ARC-c)~\citep{clark2018think}, HellaSwag (HelS)~\citep{zellers2019hellaswag}, and WinoGrande (WinG)~\citep{sakaguchi2021winogrande}. We compare our methods against several LLaMA-based ternary LLM baselines, including TWN~\cite{li2016ternary}, Spectra~\citep{kaushal2025surprising}, BitNet~\citep{ma2024era}, TernaryLLM~\citep{chen2024ternaryllm}, LLM-QAT~\citep{liu2024llm}, and ParetoQ~\citep{liu2025paretoq}.

\begin{table*}[!t]
\centering
\begin{tabular}{l|ll|lllllll}
\hline
\hline Model & Size & Bit-width & ARC-e & ARC-c & HelS & PIQA & WinG & Average\\\hline
 \rowcolor{gray!20}  LLaMA3.2 & 1B & 16 & 0.654 & 0.313 & 0.477 & 0.742 & 0.603 & 0.558\\
TernaryLLM$^*$ & 1B & 1.67 & 0.424 & 0.174 & 0.256 & 0.563 & 0.513 & 0.386\\
ParetoQ$^*$ & 1B & 1.67 & 0.421 & 0.180 & 0.273 & 0.604 & 0.510 & 0.398\\
LLM-QAT & 1B & 1.67 & 0.360 & 0.262 & 0.313 & 0.551 & 0.496 & 0.397\\
BitNet & 1.3B & 1.67 & 0.549 & 0.242 & 0.377 & 0.688 & 0.558 & 0.483\\
Spectra & 1.1B & 1.67 & 0.563 & 0.246 & 0.388 & 0.693 & 0.555 & 0.489\\
\rowcolor{HYLightBlue} Tequila & 1B & 1.67 & 0.645 & 0.305 & 0.391 & 0.710 & 0.542 & 0.519 \\
\rowcolor{HYLightBlue}Sherry & 1B & 1.25 & 0.647 & 0.309 & 0.388 & 0.699 & 0.550 & 0.519 \\ \hline \hline 
 \rowcolor{gray!20} LLaMA3.2 & 3B & 16 & 0.745 & 0.422 & 0.552 & 0.768 & 0.691 & 0.636 \\
TernaryLLM$^*$ & 3B & 1.67 & 0.361 & 0.161 & 0.260 & 0.572 & 0.496 & 0.370 \\
ParetoQ$^*$ & 3B & 1.67 & 0.498 & 0.231 & 0.303 & 0.645 & 0.529 & 0.441\\
LLM-QAT & 3B & 1.67 & 0.445 & 0.307 & 0.434 & 0.627 & 0.506 & 0.464\\
BitNet & 3B & 1.67 & 0.614 & 0.283 & 0.429 & 0.715 & 0.593 & 0.527 \\
Spectra & 3.9B & 1.67 & 0.660 & 0.319 & 0.483 & 0.744 & 0.631 & 0.567\\
\rowcolor{HYLightBlue} Tequila & 3B & 1.67 & 0.702 & 0.346 & 0.464 & 0.739 & 0.627 & 0.576 \\
\rowcolor{HYLightBlue}Sherry & 3B & 1.25 & 0.688 & 0.364 & 0.452 & 0.736 & 0.593 & 0.567\\ \hline
\end{tabular}
\vspace{0.3em}
\caption{Comparison of Tequila  and Sherry with LLaMA-based ternary LLMs across different model sizes; $^*$ indicates results obtained from our reproduction.}
\label{tb:main_result}
\end{table*}

\paragraph{Performance of Tequila and Sherry} 
The benchmark results in Table~\ref{tb:main_result} demonstrate that both Tequila and Sherry significantly outperform existing ternary quantization baselines. For the 1B model series, while traditional ternary methods suffer from a catastrophic performance drop of over 16\% on average, Tequila and Sherry reduce this gap to a marginal 3.9\% relative to the BF16 LLaMA-3.2 1B baseline. In the 3B model series, Tequila trails the full-precision model by only 6.0\%. Remarkably, Sherry achieves a 6.\% gap despite utilizing a significantly lower bit-width of 1.25 bits. When compared to other high-performance ternary baselines like BitNet, SherryLLM achieves a 4\% higher average accuracy while using 25\% fewer bits. This demonstrates that our 3:4 structured sparsification and Arenas mechanism effectively mitigate representational collapse, allowing 1.25-bit models to match or exceed the cognitive capabilities of 1.58-bit counterparts.

\paragraph{Efficiency of Tequila and Sherry} 
\begin{wraptable}{r}{0.6\textwidth} 
\centering
\begin{tabular}{l|cc|cc}\hline
Scale & Method & Bits & Speed (t/s) & Size (MB) \\ \hline
\multirow{4}{*}{0.7B} & BF16 & 16 & 34.01 & 1360.0 \\
 & BitNet(I2\_S) & 2.0 & 132.13 & 256.56 \\
 & Tequila(TL2) & 1.67 & 116.83 & 233.44 \\
 & Sherry & 1.25 & 148.27 & 205.50 \\ \hline
\multirow{4}{*}{3B} & BF16 & 16 & 7.55 & 6190.0 \\
 & BitNet(I2\_S) & 2.0 & 41.87 & 873.65 \\
 & Tequila(TL2) & 1.67 & 38.80 & 846.01 \\
 & Sherry & 1.25 & 45.55 & 712.40 \\ \hline
\end{tabular}
\vspace{0.3em}
\caption{Inference efficiency comparison on Intel i7-14700HX.}
\label{tab:inference_speed}
\end{wraptable}
We evaluate the hardware efficiency on an Intel i7-14700HX CPU, comparing throughput (tokens per second) and model size as shown in Table~\ref{tab:inference_speed}. Sherry consistently exhibits the highest efficiency across all scales. By restoring power-of-two alignment through its 5-bit packing strategy, Sherry achieves a generation speed of 148.27 t/s for the 0.7B scale, surpassing BitNet (I\_S, 2.0 bits) and Tequila (TL2, 1.67 bits). Furthermore, Sherry reduces the model size by approximately 20\% compared to 2-bit counterparts. These results confirm that AngelSlim's hardware-aligned sparsification provides a superior "sweet spot" for deploying high-performance LLMs on edge devices with limited memory bandwidth.

\subsection{Post-training Quantization}

Post-training Quantization (PTQ) serves as a critical enabling technology for deploying large-scale neural networks efficiently, as it transforms high-precision models into compact, inference-optimized forms without requiring retraining. In this section, we will first introduce the PTQ framework of AngelSlim, and then present our proposed LeptoQuant.

\subsubsection{PTQ Framework}

To address the storage and computational bottlenecks in deploying large models, Angelslim provides a unified post-training quantization (PTQ) framework. The framework not only integrates current mainstream quantization schemes but also offers a complete toolchain for analysis, ensuring the reliability of quantized models. Our AngelSlim framework is characterized by the following key features:

\begin{itemize}
\item Highly Integrated: This toolkit integrates mainstream compression algorithms into a unified framework, offering developers one-click access with exceptional ease of use.
\item Continuous Innovation: Beyond integrating widely-used industry algorithms, we are continuously researching better compression algorithms, which will be gradually open-sourced in the future.
\item Performance-Driven: We continuously optimize end-to-end performance in model compression workflows and algorithm deployment, such as enabling quantization of models like Qwen3-235B and DeepSeek-R1 on a single GPU.
\end{itemize}

\begin{figure}[t]
    \centering
    \includegraphics[width=0.9\linewidth]{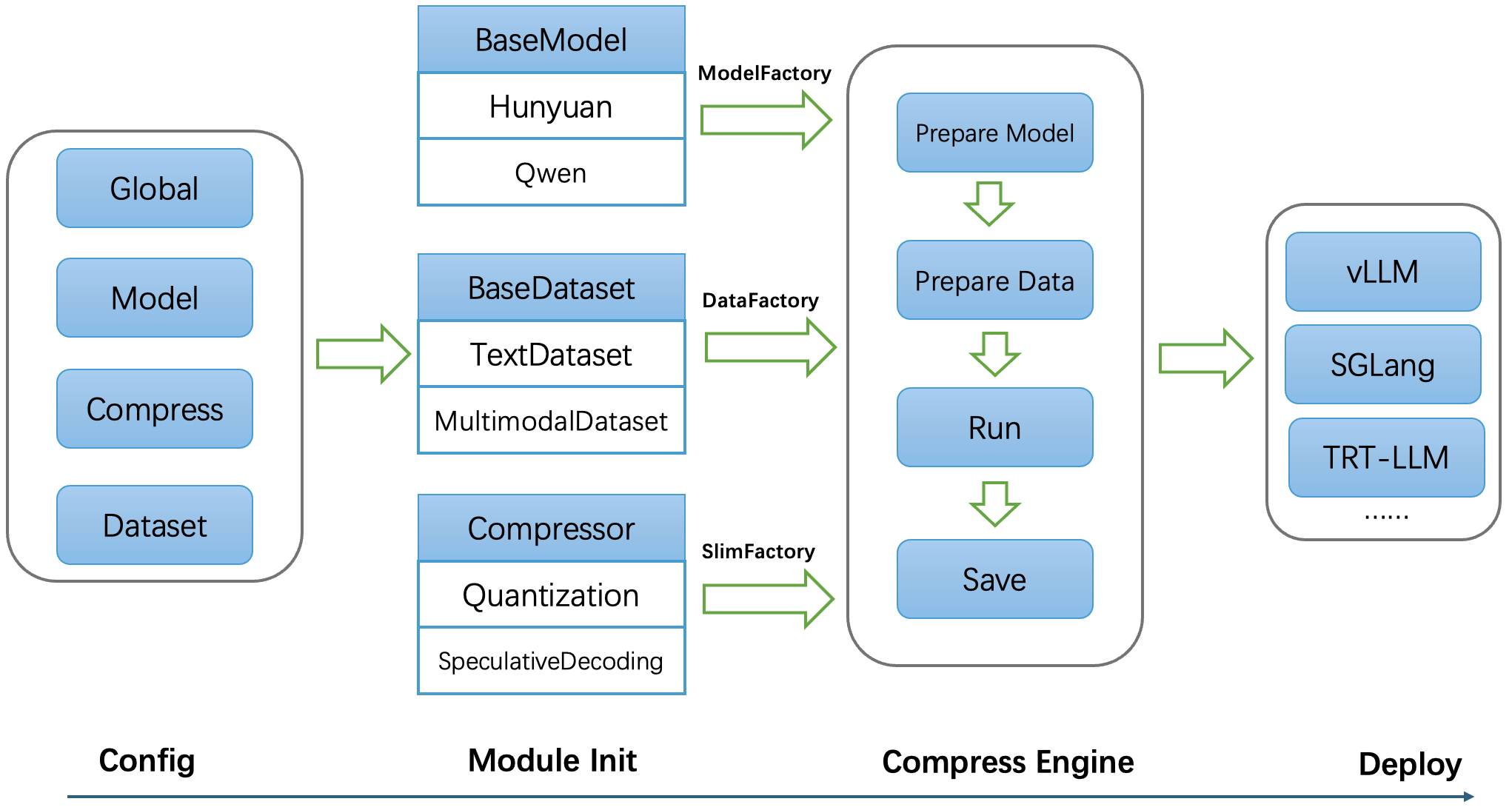}
    \caption{AngelSlim Quantization Framework.}
    \label{fig:angelslim_quant}
\end{figure}

As shown in Figure~\ref{fig:angelslim_quant}, AngelSlim starts by parsing a YAML configuration file to load all essential parameters for the compression task. This includes global settings, model information, compression algorithm specifications, and dataset configurations. This centralized configuration approach enables users to customize the pipeline without modifying the core code, ensuring high flexibility and reproducibility.

Moving to the Module Init stage, the framework initializes three specialized factories to encapsulate its core components: the ModelFactory manages the registration and instantiation of base models such as Hunyuan~\cite{hunyuana13b} and Qwen~\cite{qwen3}, allowing seamless integration of new LLMs through a registration mechanism; the DataFactory, built on a DataLoaderFactory, encapsulates diverse dataset types including TextDataset and MultimodalDataset to create unified data loaders, ensuring consistent data preprocessing across tasks; and the SlimFactory provides access to a suite of compression techniques such as quantization and speculative decoding, which are registered and dispatched dynamically based on the configuration to support multiple compression strategies in a unified manner.

The Compress Engine stage then orchestrates the end-to-end compression workflow: it first prepares the target model from the ModelFactory and initializes the data loader from the DataFactory, then executes the selected compression algorithm from the SlimFactory to reduce model size and latency while preserving performance, before saving the compressed model checkpoint for deployment. Finally, the compressed models can be directly deployed via high-performance inference backends, including vLLM~\cite{kwon2023efficient} and SGLang~\cite{zheng2024sglang}. Together, these stages form a streamlined, extensible pipeline that unifies model compression and deployment, making it adaptable to a wide range of LLM optimization scenarios.

At the algorithmic level, the framework covers the full precision spectrum from FP8, INT8 to INT4. FP8 quantization focuses on achieving significant speed-up with limited precision loss, supporting both dynamic and static modes: Dynamic Quantization (W8A8-FP8 Dynamic) performs real-time scaling on activation values to adapt to input variations; whereas Static Quantization (W8A8-FP8 Static) determines a unified scaling factor via offline calibration, pursuing ultimate inference speed. For extremely large models, its unique Low-Memory FP8 Calibration mode effectively reduces the GPU memory overhead of the quantization process itself. For higher compression demands, the framework provides two mainstream INT4 weight quantization schemes: The AWQ~\cite{lin2024awq} algorithm adaptively amplifies the numerical range of important weight channels by analyzing activation distributions, thereby preserving more information under 4-bit quantization; the GPTQ algorithm employs layer-wise reconstruction optimization to minimize the output error after quantization. Both require only a small amount of calibration data, operate without training, and achieve a balance between accuracy and compression efficiency.

To ensure quantization quality, the framework includes built-in diagnostic analysis tools. Users can enable Scale Analysis during quantization to detect outliers, and after quantization, use dedicated tools to compare the distribution differences between FP8 and the original BF16 weights, providing data insights for potential accuracy restoration.

The framework's practicality is reflected in its broad adaptation to mainstream model families. For the Hunyuan series models, it primarily supports FP8, INT4-AWQ, and INT4-GPTQ schemes. For the DeepSeek-R1~\cite{liu2024deepseek} model, a more aggressive mixed-precision strategy is further implemented: its W4A8-FP8 scheme performs group-wise INT4 quantization on weights (group size 128) while keeping activation values in FP8 precision, achieving in-depth optimization of storage and computational efficiency; its INT4-AWQ scheme can also seamlessly integrate with inference engines like vLLM. Additionally, the framework is compatible with diverse Qwen series models, including multimodal models, demonstrating good generalization capability.

In addition, the Angelslim framework takes into account the hardware resource limitations in practical deployment. Its Low-Memory calibration mode implements an intelligent CPU-offloading strategy: instead of keeping all intermediate layer parameters in GPU memory, it strategically stores them in the much larger CPU RAM, loading them back to the GPU only when computation is required. This on-demand swapping mechanism significantly reduces the peak GPU memory overhead of the quantization process itself while ensuring calibration accuracy. This enables large-scale models like DeepSeek-R1 to complete the entire calibration process using only a single GPU, without relying on multi-GPU or high-memory configurations.

\begin{table}[t]
\centering
\resizebox{\textwidth}{!}{%
\begin{tabular}{lccccc}
\toprule
\textbf{Model} & \textbf{Quantization} & \textbf{GPQA Diamond} & \textbf{AIME 2024} & \textbf{SimpleQA} & \textbf{LiveCodeBench}\\ \midrule
\multirow{2}{*}{DeepSeek-R1-0528} & FP8-Block-Wise & 78.28 & 88.67 & 27.8 & 77.1  \\
 & W4A8-FP8 & 77.37 & 88.67 & 26.83 & 78.86 \\ \bottomrule
\end{tabular}
}
\vspace{0.3em}
\caption{Benchmark results for DeepSeek-R1-0528 model with FP8-Block-Wise and W4A8-FP8 quantization algorithms on datasets including GPQA Diamond, AIME 2024, SimpleQA and LiveCodeBench}
\label{tab:deepseekr1}
\vspace{-1em}
\end{table}

Beyond memory efficiency, the framework's core strength lies in its ability to preserve model accuracy under aggressive quantization.  Table~\ref{tab:deepseekr1} demonstrates its efficacy, showcasing near-lossless accuracy retention for DeepSeek-R1 even under highly compressed W4A8 schemes, which is a significant challenge given the model's scale and reasoning capabilities.

In summary, by constructing a complete ecosystem encompassing multiple algorithms, granularities, and models, and equipped with analysis tools, the Angelslim PTQ framework provides a practical, engineering-ready solution for the efficient deployment of large-scale language models.

\subsubsection{LeptoQuant}

For PTQ (Post-Training Quantization), In this section, we analyzed the numerical precision loss experienced by models after PTQ quantization to FP8. Then we introduced the Dynamic Outlier Isolation Scale (LeptoQuant), which addresses this anomalous loss by searching for optimal scaling values to mitigate quantization loss. The figure shows the overall architecture. 

Typically, PTQ calculates the abs Max values of activations and weights as quantization scaling factors. By observing the numerical distribution of models after FP8 PTQ quantization, we found that the variance of activation distributions is significantly greater than that of bf16 models. This numerical distribution causes quantized values to fall within a range that is difficult to represent in FP8, resulting in excessive losses on mathematically difficult tasks or those requiring strict text formatting.

\begin{figure*}[t]
    \centering
     \includegraphics[width=0.8\textwidth]{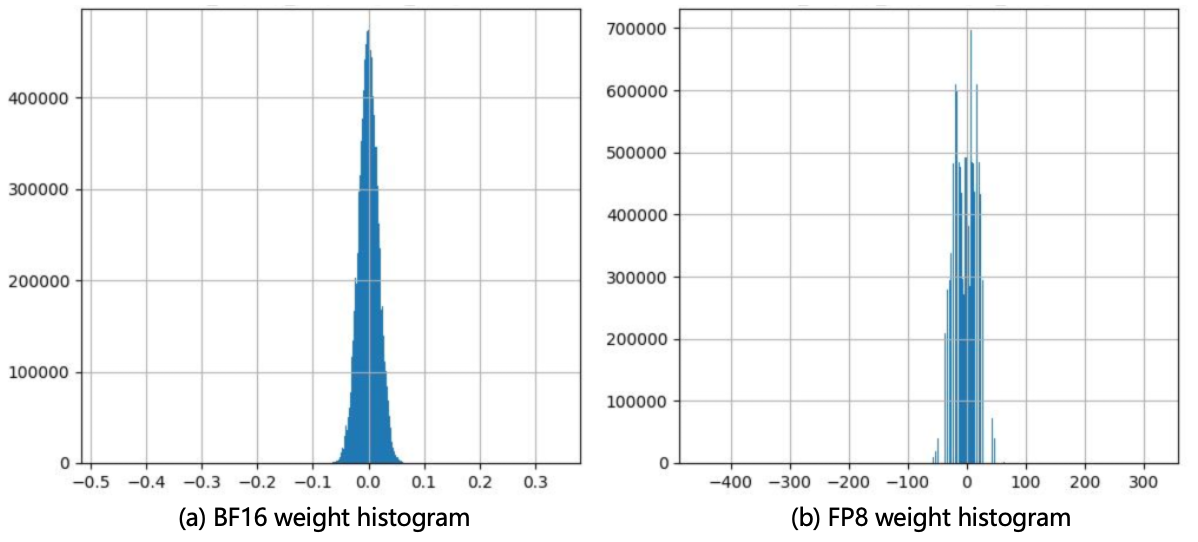}
    \caption{Figure $a$ is the BF16 weight distribution histogram of hunyuan-4b-insturct, and Figure $b$ is the weight distribution histogram after FP8 quantization. It can be seen that the weight distribution after quantization is slightly smoothed to a position far away from 0.
    } 
    \label{fig:anaylse_bf16_fp8_4b.png}
    \vspace{-1em}
\end{figure*}

Figure~\ref{fig:anaylse_bf16_fp8_4b.png} shows a histogram of the numerical distribution of the last layer of $FFN2$ before and after quantization in a model with significant FP8 quantization loss. By observing the original precision numerical distribution, it is found that the overall numerical distribution of the weight is concentrated in a peak distribution, with obvious outliers and most of the data concentrated near zero. The relative distance between the values is small and the data distribution approaches the laplace distribution, resulting in higher precision requirements during the calculation process. The weight distribution after FP8 QDQ is shown in the figure on the right. It can be seen that the quantized distribution is smoother than the original precision. Since the numerical expression of FP8-E4M3 is closer to 0, the more values can be represented, and the original precision weight under the normal distribution is close to Worigin. Traditional FP8 quantization causes the originally densely populated numbers to be smoothed to the FP8 precision range with poor expressiveness, resulting in reduced precision expression and loss of performance.

To address the aforementioned FP8 quantization issues, we introduced Leptokurtic Quant (LeptoQuant) search, a search strategy that concentrates the FP8 weight mapping range into a high-precision region by isolating outliers. Quantizing activations is generally more difficult than quantizing weights, so we prioritize optimizing FP8 activations. LeptoQuant uses the original FP8 outlier values as the upper limit of FP8 precision expression, calculating a new scale that compresses the value distribution into a high-precision distribution range, resulting in better precision expression of quantized activation values.

\textbf{Searching to scale.} Specifically, we define the FP8 value obtained by traditional PTQ calculation as ${{FP8}_{origin}}$, and then introduce a hyperparameter $\alpha$, which determines the proportion of isolated outliers. Generally speaking, the quantization of activation values is much more difficult than weights, so we will prioritize optimize the activation. Our search range is from $M$ to 0. For a given $\alpha$, we calculate the outlier value of ${{FP8}_{origin}}$ as the new scaling factor denominator $D$, thereby constraining the FP8 quantization expression range and compressing the values in the densely distributed area to a position with higher precision. The expression of $D$ is as follows:
\begin{equation}
\label{eq_D}
D_w=Outlier(W, {\alpha})
, \quad  W=\max |FP8_{origin}|,
\end{equation}
where $\alpha$ is selected to start searching from the 0.1\% outlier, $Outlier$ represents the value of position a in $W$ given the weight and $\alpha$.

\textbf{Calculating the quantization error.} We calculate the loss of each block layer by layer through dynamic interpolation. Consider a $FFN$ block, we obtain the activation and weight of the model, and the calculation of the $FFN$ block can be written as $O=block(x)$. Then we introduce $formula $ into the block forward reasoning process to simulate the actual FP8 quantization calculation results of $LeptoQuant$. Specifically, the formula for quantifying $FFN1$ can be defined as:
\begin{equation}
\label{eq_Y}
\hat{Y}_D=Block(w,x,Scale_D) , \quad Scale_D=\frac{w_{max} }{D_{w}},
\end{equation}
where $D_{w}$ is the denominator of the scaling factor of activation $w$, $Scale_D$ is the FP8 quantization scaling factor, and $\hat{Y}_D$ is the block-wise calculation result after quantizing the activation to FP8. Therefore, for each OP, we choose to automatically search for an optimal scaling factor to minimize the output error after quantization of a specific block. Formally, we want to optimize the following objective:
\begin{equation}
\label{eq_L}
L(s)=\frac{1}{n} \sum_{i=1}^n (Y_i - \hat{Y}_i)^2, \quad \{ i \mid i \in \mathbb{Z}, 0 \leq i \leq n_{samlps} \},
\end{equation}
where $i$ indicates different samples. For the search process, we use a more stable search space,
fast grid search to select $\alpha$ in the interval $[0, 0.001]$ (0 means traditional FP8 quantization without isolating outliers, and 0.001 means the most aggressive scaling in the search space) and calculate the optimal $\alpha$ using the MSEloss.

\paragraph{LeptoQuant Benchmark}

For open-source models, we conducted LeptoQuant experiments on models and datasets from the Hunyuan and Qwen series that showed significant performance degradation after FP8 quantization, as shown in the Table. ~\ref{tab:LeptoQuant_Benchmark_QWEN} and ~\ref{tab:LeptoQuant_Benchmark_HY} .

\begin{table}[t]
\centering
\begin{tabular}{c|cccc}
\hline
Base Model & Type & OlympiadBench & AIME\_2024 & AIME\_2025 \\ 
\hline
\multirow{3}{*}{Hunyuan-4B-Instruct} 
& BF16 & 73.10 & 78.30 & 66.50 \\ 
& FP8 & 72.40 & 66.70 & 46.70 \\ 
& FP8-lepto & 73.10 & 76.66 & 60.70 \\ 
\hline
\multirow{3}{*}{Hunyuan-2B-Instruct} 
& BF16 & 73.41 & 56.67 & 53.85 \\ 
& FP8 & 72.41 & 50.00 & 37.00 \\ 
& FP8-lepto & 72.64 & 53.00 & 36.00 \\ 
\hline
\end{tabular}

\vspace{0.3em}
\caption{Compare the performance of the standard FP8 with that of the HY series models using leptoQuant}
\label{tab:LeptoQuant_Benchmark_HY}
\end{table}

\begin{table}[t]
\begin{tabular}{c|cccc}
\hline
Base Model & Type & GSM8K-flexible-extract & GSM8K-strict-match & HUMANEVAL \\ \hline
\multirow{3}{*}{Qwen2.5-1.5B-Instruct} & BF16      & 59.00                   & 54.20               & 37.20      \\
                                       & FP8       & 55.72                  & 49.36              & 32.93     \\
                                       & FP8-lepto & 57.09                  & 52.31              & 35.37     \\ \hline
\multirow{3}{*}{QWEN3-0.6B}            & BF16      & 41.62                  & 41.7               & 20.12     \\
                                       & FP8       & 37.83                  & 38.06              & 20.73     \\
                                       &  FP8-lepto & 38.59                  & 38.44              & 21.95     \\ \hline
\multirow{3}{*}{QWEN3-4B}              & BF16      & 84.91                  & 85.44              & 73.17     \\
                                       & FP8       & 83.85                  & 83.85              & 67.7      \\
                                       &  FP8-lepto & 83.62                  & 83.93              & 68.9      \\ \hline
\end{tabular}
\vspace{0.3em}
\caption{Compare the performance of the standard FP8 with that of the Qwen series models using leptoQuant}
\label{tab:LeptoQuant_Benchmark_QWEN}
\end{table}

\section{Speculative Decoding}

\subsection{Training Framework}
Speculative decoding has emerged as a critical technique for accelerating LLM inference; however, the training of draft models remains poorly supported by existing frameworks. Unlike standard language modeling, draft model training is inherently \emph{target-model–dependent}: the objective is not to optimize standalone generation quality, but to maximize alignment with the token distribution and generation dynamics of a fixed target model. This dependency introduces requirements—such as target-model feedback, multi-step prediction supervision, and customized execution flows—that are difficult to express within conventional training pipelines.

These challenges are further amplified by advanced methods such as Eagle-3. Its tree-based attention structure and multi-step verification impose strict constraints on data ordering, attention masking, and temporal consistency during training. Meanwhile, the rapid evolution of large-scale architectures—including Mixture-of-Experts (MoE) and long-context Transformers—demands training solutions that are both scalable and compatible with production inference engines. Existing approaches typically address these aspects in isolation, resulting in fragmented systems that are difficult to maintain, extend, or deploy.

To address these limitations, we design a dedicated speculative sampling training framework in AngelSlim, with Eagle-3 compatibility, production inference integration, and architectural scalability as first-class design goals.

\begin{figure}[t]
    \centering
    \includegraphics[width=0.95\linewidth]{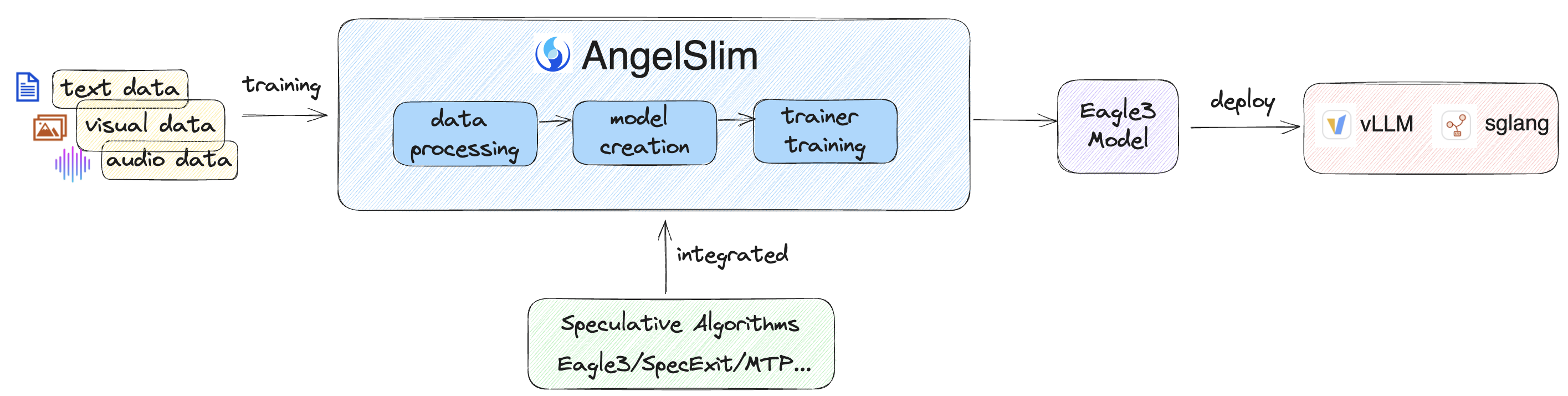}
    \caption{AngelSlim speculative training framework.
    The framework unifies data processing, model construction,
    and trainer-level execution to support
    \emph{target-model–dependent} draft model training.}
    \label{fig:speculative-training-framework}
\end{figure}

\subsubsection{Overview and Objective}
AngelSlim is a unified model compression toolkit that integrates multiple acceleration techniques—including quantization and speculative decoding—across language, vision--language, and speech models. In this section, we focus on its speculative sampling training framework, which treats speculative decoding as a \emph{first-class training objective} rather than a post-hoc inference optimization.

Speculative sampling accelerates inference by decoupling generation and verification: a lightweight draft model predicts multiple future tokens in parallel, while a target model verifies these predictions in a single step. AngelSlim operationalizes this paradigm through an end-to-end training pipeline—from data preparation and model abstraction to algorithm-specific execution—explicitly optimizing draft models to align with the target model’s token distribution and generation dynamics.

By design, the framework supports Eagle-3–style training while remaining deployment-oriented and modality-agnostic, enabling draft models trained in AngelSlim to be directly integrated into production inference engines.

\subsubsection{Design Principles}
AngelSlim’s speculative sampling framework is built around three core principles:

\textbf{Eagle-3 train-and-deploy compatibility.}
Draft models trained in AngelSlim can be directly deployed through the speculative decoding interfaces of production inference engines such as vLLM and SGLang, without additional conversion or parameter tuning.

\textbf{Multi-modality by design.}
Text, vision--language, and speech models share a unified set of training abstractions and algorithmic components, avoiding modality-specific reimplementation.

\textbf{Scalability and extensibility.}
The framework natively supports large-scale architectures, including MoE models and long-context Transformers, enabling efficient training across a wide range of model sizes.

Across diverse models and modalities, AngelSlim-trained Eagle-3 draft models achieve a consistent \textbf{1.4$\times$–1.9$\times$ inference speedup} under standard speculative decoding settings.

\subsubsection{Core Training Components}

\paragraph{Data Processing Module}
The data pipeline provides a reusable and stable foundation for speculative training across modalities. It includes:
\begin{itemize}[left=1.5em, topsep=0pt, partopsep=0pt, parsep=0pt, itemsep=2pt]
    \item Data resampling to construct in-distribution training data aligned with the target model.
    \item Unified preprocessing that converts text, image, and audio inputs into standardized token IDs and loss masks, including vocabulary mapping for pruned draft models.
    \item Hidden state extraction from the target model, serving as supervision signals for draft model training.
\end{itemize}

\begin{table}[t]
\centering
\resizebox{\textwidth}{!}{
\begin{tabular}{l l r r r r r r r r r r}
\toprule
\multirow{2}{*}{\textbf{Model}} & \multirow{2}{*}{\textbf{Method}} & \multicolumn{2}{c}{\textbf{GSM8K}} & \multicolumn{2}{c}{\textbf{Alpaca}} & \multicolumn{2}{c}{\textbf{HumanEval}} & \multicolumn{2}{c}{\textbf{MT-bench}} & \multicolumn{2}{c}{\textbf{Mean}} \\
\cmidrule(lr){3-4}\cmidrule(lr){5-6}\cmidrule(lr){7-8}\cmidrule(lr){9-10}\cmidrule(lr){11-12}
 & & \textbf{TPS} & \textbf{AL} & \textbf{TPS} & \textbf{AL} & \textbf{TPS} & \textbf{AL} & \textbf{TPS} & \textbf{AL} & \textbf{TPS} & \textbf{AL} \\
\midrule
\multirow{2}{*}{\textbf{Qwen3-1.7B}} & Vanilla & 376.42 & 1 & 378.86 & 1 & 378.38 & 1 & 390.53 & 1 & 381.05 & 1 \\
 & \href{https://huggingface.co/AngelSlim/Qwen3-1.7B_eagle3}{Eagle3} & 616.9 & 2.13 & 653.29 & 2.19 & 680.1 & 2.2 & 621.44 & 2.17 & 642.93 & 2.17 \\
\midrule
\multirow{2}{*}{\textbf{Qwen3-4B}} & Vanilla & 229.05 & 1 & 235.29 & 1 & 234.66 & 1 & 234.04 & 1 & 233.26 & 1 \\
 & \href{https://huggingface.co/AngelSlim/Qwen3-4B_eagle3}{Eagle3} & 389.35 & 2.07 & 395.97 & 2.1 & 377.84 & 2.08 & 384.6 & 2.07 & 386.94 & 2.08 \\
\midrule
\multirow{2}{*}{\textbf{Qwen3-8B}} & Vanilla & 149.63 & 1 & 149.93 & 1 & 153.85 & 1 & 153.81 & 1 & 151.81 & 1 \\
 & \href{https://huggingface.co/AngelSlim/Qwen3-8B_eagle3}{Eagle3} & 257.32 & 2 & 266.69 & 2.02 & 244.89 & 1.97 & 258.2 & 1.97 & 257.52 & 1.99 \\
\midrule
\multirow{2}{*}{\textbf{Qwen3-14B}} & Vanilla & 92.97 & 1 & 92.66 & 1 & 92.94 & 1 & 94.46 & 1 & 93.26 & 1 \\
 & \href{https://huggingface.co/AngelSlim/Qwen3-14B_eagle3}{Eagle3} & 153.72 & 1.87 & 140.46 & 1.78 & 144.68 & 1.76 & 142.45 & 1.74 & 145.33 & 1.79 \\
\midrule
\multirow{2}{*}{\textbf{Qwen3-32B}} & Vanilla & 43.49 & 1 & 43.38 & 1 & 43.19 & 1 & 43.3 & 1 & 43.32 & 1 \\
 & \href{https://huggingface.co/AngelSlim/Qwen3-32B_eagle3}{Eagle3} & 80.43 & 2.01 & 72.49 & 1.9 & 71.57 & 1.86 & 74.1 & 1.86 & 74.1 & 1.91 \\
\midrule
\multirow{2}{*}{\textbf{Qwen3-30B-A3B}} & Vanilla & 311.84 & 1 & 320.43 & 1 & 325.77 & 1 & 325.42 & 1 & 320.87 & 1 \\
 & \href{https://huggingface.co/AngelSlim/Qwen3-a3B_eagle3}{Eagle3} & 453.97 & 2.1 & 432.45 & 2.04 & 428.81 & 2.02 & 437.06 & 2.01 & 438.07 & 2.04 \\
\bottomrule
\end{tabular}
}
\vspace{2mm}
\caption[Qwen3 series benchmark]{%
Qwen3 series models benchmark using Eagle3 speculative decoding on vLLM (tp=1, ep=1, num\_speculative\_tokens=2, batch\_size=1, output\_len=1024). \textit{TPS: throughput (tokens/s); AL: average accepted speculative tokens per decoding step.}}
\label{tab:qwen3_series_eagle3}
\end{table}

\begin{table*}[t]
\centering
\resizebox{\textwidth}{!}{
\begin{tabular}{llrrrrrrrr} %
\toprule
\multirow{2}{*}{\textbf{Model}} & \multirow{2}{*}{\textbf{Method}}
& \multicolumn{2}{c}{\textbf{MT-Bench}}
& \multicolumn{2}{c}{\textbf{MATH-500}}
& \multicolumn{2}{c}{\textbf{MMMU}}
& \multicolumn{2}{c}{\textbf{MMStar}} \\
\cmidrule(lr){3-4}\cmidrule(lr){5-6}\cmidrule(lr){7-8}\cmidrule(lr){9-10}
 &  & \textbf{TPS} & \textbf{AL}
 & \textbf{TPS} & \textbf{AL}
 & \textbf{TPS} & \textbf{AL}
 & \textbf{TPS} & \textbf{AL} \\
\midrule

\multirow{2}{*}{\textbf{Qwen3-VL-2B-Instruct}}
& Vanilla
& 346.31 & 1
& 82.96  & 1
& 83.27  & 1
& 81.63  & 1 \\
& \href{https://huggingface.co/AngelSlim/Qwen3-VL-2B-Instruct_eagle3}{Eagle3}
& 555.22 & 2.29
& 163.09 & 2.57
& 154.18 & 2.55
& 139.73 & 2.31 \\

\midrule
\multirow{2}{*}{\textbf{Qwen3-VL-4B-Instruct}}
& Vanilla
& 212.10 & 1
& 67.96  & 1
& 65.88  & 1
& 67.75  & 1 \\
& \href{https://huggingface.co/AngelSlim/Qwen3-VL-4B-Instruct_eagle3}{Eagle3}
& 382.33 & 2.34
& 141.87 & 2.72
& 104.44 & 2.05
& 107.07 & 2.10 \\

\midrule
\multirow{2}{*}{\textbf{Qwen3-VL-30B-A3B-Instruct}}
& Vanilla
& 180.57 & 1
& 31.08  & 1
& 31.51  & 1
& 30.93  & 1 \\
& \href{https://huggingface.co/AngelSlim/Qwen3-VL-30B-A3B-Instruct_eagle3}{Eagle3}
& 240.47 & 2.19
& 75.31  & 2.79
& 48.47  & 1.78
& 52.57  & 1.94 \\

\bottomrule
\end{tabular}
}
\vspace{1mm}
\caption[Qwen3-VL series benchmark]{%
Qwen3-VL series models benchmark using Eagle3 speculative decoding on vLLM (tp=1, ep=1, num\_speculative\_tokens=4, batch\_size=1, output\_len=1024).
\textit{TPS: throughput (tokens/s); AL: average accepted speculative tokens per decoding step.}}
\label{tab:qwen3_vl_eagle3}
\end{table*}

\paragraph{Model Abstraction Module}
Model extensibility is enabled through a unified \texttt{TargetModel} interface, which abstracts:
\begin{itemize}[left=1.5em, topsep=0pt, partopsep=0pt, parsep=0pt, itemsep=2pt]
    \item Model loading and weight management
    \item Forward execution
    \item Intermediate and hidden state access
\end{itemize}

To support a new architecture or backend, users only need to implement the required abstract methods, without modifying the trainer or algorithm logic. This design significantly reduces the engineering overhead required to adapt AngelSlim to new models and modalities.

\paragraph{Trainer Module}
The trainer encapsulates the algorithm-specific logic of Eagle-3 and supports two execution modes:
\begin{itemize}[left=1.5em, topsep=0pt, itemsep=2pt]
    \item \textbf{Online training}, where hidden states are computed on the fly, suitable for smaller models or memory-rich environments.
    \item \textbf{Offline training}, where hidden states are precomputed and stored, enabling efficient training of large models under limited GPU memory.
\end{itemize}

To ensure stability and correctness during training, the trainer further provides:
\begin{itemize}[left=1.5em, topsep=0pt, itemsep=2pt]
    \item \textbf{Training-time testing}, which explicitly simulates multi-step speculative generation so that the draft model learns to condition on its own predictions.
    \item \textbf{Native checkpointing support}, enabling full recovery of model parameters, optimizer states, and training progress.
\end{itemize}

These safeguards ensure that the resulting draft models behave robustly under speculative decoding at inference time. In practice, AngelSlim-trained models can be directly deployed in vLLM-based inference pipelines, consistently improving acceptance length and end-to-end throughput across diverse tasks.

\subsubsection{End-to-End Performance}
We evaluate end-to-end performance across a diverse set of benchmarks, covering language-only, vision--language, OCR, and speech tasks. All experiments employ Eagle-3 speculative decoding on vLLM with tp=1, ep=1, and task-specific numbers of speculative tokens.

For the Qwen3 series (Table~\ref{tab:qwen3_series_eagle3}), Eagle-3 consistently improves throughput (TPS) across all model sizes relative to vanilla decoding. The average accepted speculative length (AL) ranges from 1.74 to 2.2, indicating effective multi-token speculation while preserving correctness. Larger models exhibit particularly strong speedups, suggesting favorable scaling behavior.

In multimodal settings, the Qwen3-VL series (Table~\ref{tab:qwen3_vl_eagle3}) similarly benefits from Eagle-3 across MT-Bench, MATH-500, MMMU, and MMStar. Smaller models achieve AL values above 2.3, while the largest model shows slightly reduced AL on certain benchmarks, reflecting the trade-off between speculative depth and verification cost at scale.

Non-language tasks also demonstrate substantial gains. Hunyuan-OCR on OmniDocBench (Table~\ref{tab:eagle3_benchmark_all}) improves throughput from 70.12 to 108.1 TPS with an AL of 2.08. Speech models, including Qwen2-Audio on LibriSpeech and Fun-CosyVoice3 on LibriTTS, similarly achieve significant throughput improvements while maintaining high acceptance rates.

Overall, Eagle-3 speculative decoding delivers a consistent \textbf{1.8$\times$–2.0$\times$ end-to-end speedup} across all AngelSlim-trained models, demonstrating its effectiveness as a scalable and reliable acceleration technique for both language and multimodal inference.

\begin{table*}[t]
\centering
  \setlength{\tabcolsep}{5pt}
  \begin{tabular}{l l l r r}
    \toprule
    \textbf{Model} & \textbf{Dataset} & \textbf{Method} & \textbf{TPS} & \textbf{AL} \\
    \midrule
    \multirow{2}{*}{\textbf{Hunyuan-OCR}} & \multirow{2}{*}{OmniDocBench} & Vanilla & 70.12 & 1 \\
    & & \href{https://huggingface.co/AngelSlim/HunyuanOCR_eagle3}{Eagle3} & 108.1 & 2.08 \\
    \midrule
    \multirow{2}{*}{\textbf{Qwen2-Audio}} & \multirow{2}{*}{LibriSpeech} & Vanilla & 78.76 & 1 \\
    & & \href{https://huggingface.co/AngelSlim/Qwen2-Audio-7B-Instruct_eagle3}{Eagle3} & 146.66 & 3.51 \\
    \midrule
    \multirow{2}{*}{\textbf{Fun-CosyVoice3}} & \multirow{2}{*}{LibriTTS} & Vanilla & - & 1 \\
    & & \href{https://huggingface.co/AngelSlim/Fun-CosyVoice3-0.5B-2512_eagle3}{Eagle3} & - & 1.96 \\
    \bottomrule
  \end{tabular}
\vspace{1mm}
  \caption{Benchmark results of Eagle3 speculative decoding on vLLM (Hunyuan-OCR $\&$ Qwen2-Audio $\&$ Fun-CosyVoce, tp=1, ep=1, num\_speculative\_tokens=4, batch\_size=1, output\_len=1024). \textit{TPS: throughput (tokens/s); AL: average accepted speculative tokens per decoding step.}}
  \label{tab:eagle3_benchmark_all}
\end{table*}

\subsection{Algorithmic Innovations}

\begin{figure}[!ht]
    \centering
    \includegraphics[width=0.9\linewidth]{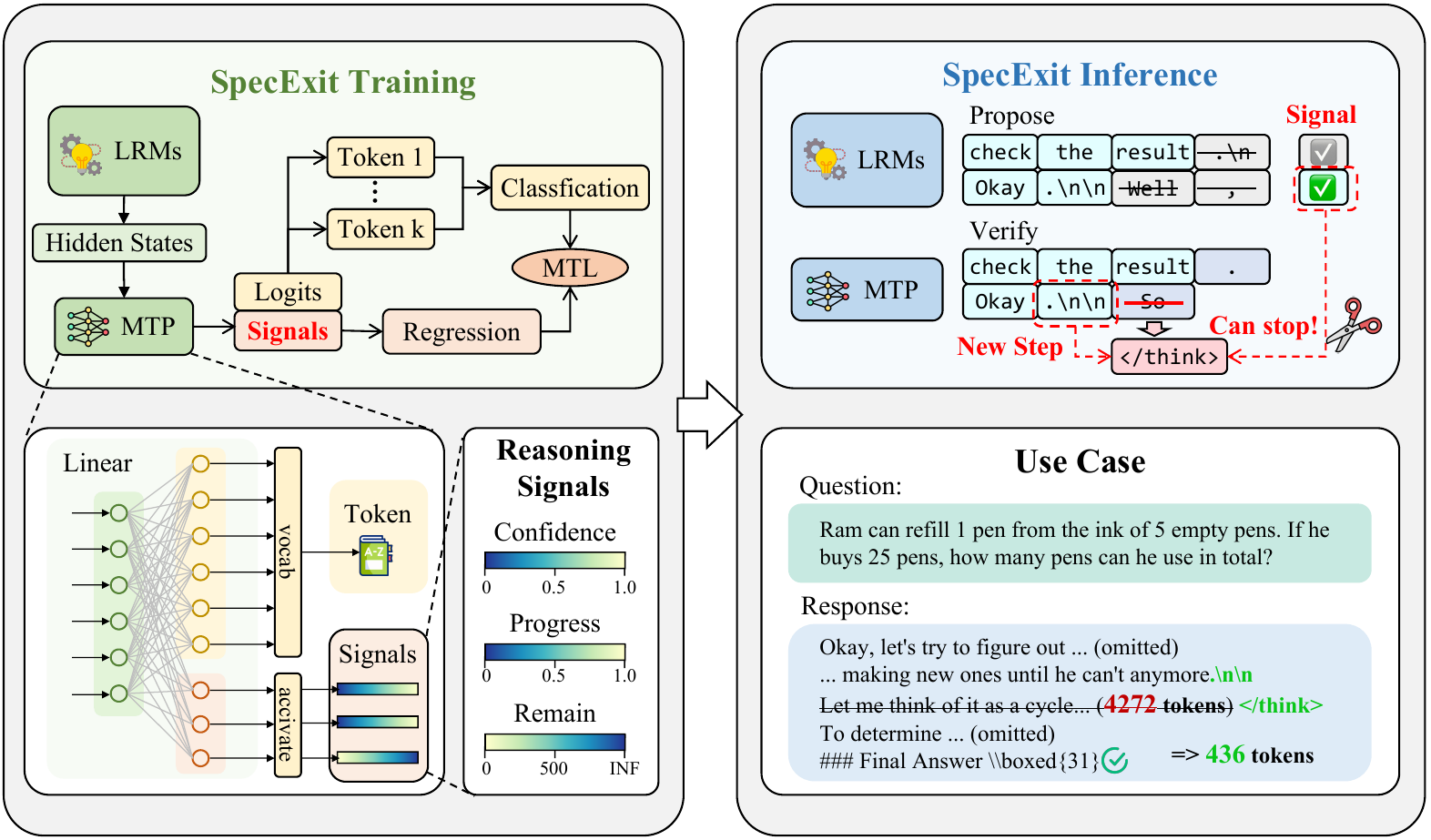}
    \caption{Overall architecture of the proposed SpecExit framework. The Multi-Token Prediction (MTP) layer is augmented to output both token logits and auxiliary signals. Training is performed with multi-task learning, while at inference these signals guide speculative early stopping without modifying the backbone model. The example illustrates how redundant reasoning steps can be pruned while preserving final answer quality.}
    \label{fig:specexit_method}
\end{figure}

In addition to the system-level and framework advancements presented earlier, we also draw inspiration from recent algorithmic innovations in the efficient acceleration of large reasoning models. One notable example is \textit{SpecExit}~\cite{yang2025specexit}, a novel approach designed to address inefficiencies caused by “overthinking” in large reasoning models (LRMs). LRMs often generate excessively long reasoning chains, which, while improving accuracy, result in high latency and unnecessary computation. Traditional early-exit mechanisms mitigate this issue by terminating generation once a sufficient answer is formed, but they typically introduce detection overhead and lack generalizability.

SpecExit overcomes these limitations by integrating early-exit prediction directly into a lightweight draft model’s hidden states. This enables the model to simultaneously propose future tokens and generate early-exit signals without additional probing overhead. The method leverages latent signals such as confidence, reasoning progress, and remaining steps to dynamically determine when to exit the reasoning process early. As a result, SpecExit can significantly reduce the average generation length—reported to decrease by approximately 66\%—and achieve substantial end-to-end latency improvements, with up to a 2.5× speedup compared to speculative decoding baselines, while maintaining task accuracy.

By exploiting hidden-state information for dynamic reasoning control, SpecExit exemplifies how algorithmic innovations can be combined with speculative decoding techniques to accelerate large model inference in practical applications.

\subsubsection{Method: SpecExit}

SpecExit extends speculative decoding by embedding early-exit decision making directly into the draft model, enabling dynamic regulation of reasoning depth without introducing additional detection modules or modifying the backbone language model. The core insight is that the hidden states used for next-token prediction also encode higher-level information about reasoning progress, confidence, and content completeness. To make these signals explicit and learnable, SpecExit augments the draft model’s Multi-Token Prediction (MTP) layer to jointly predict speculative token logits and auxiliary reasoning-related signals. In addition to projecting hidden states into the vocabulary space for multi-token prediction, the extended MTP layer includes lightweight auxiliary heads that estimate confidence, reasoning progress, and remaining reasoning length, while preserving the draft model’s original language modeling capability.

Figure~\ref{fig:specexit_method} illustrates the overall SpecExit architecture and workflow. During training, the extended MTP layer is optimized via multi-task learning, jointly supervising token prediction and auxiliary signal estimation. At inference time, the draft model simultaneously proposes speculative tokens and corresponding exit signals in a single forward pass. These signals guide early stopping decisions within the speculative decoding loop: if the predicted reasoning state indicates sufficient completeness, generation terminates early; otherwise, speculative tokens are verified by the target model as in standard speculative decoding. By integrating early-exit prediction directly into the draft–verify pipeline, SpecExit retains the low-latency benefits of speculative decoding while avoiding the overhead and brittleness of external stopping criteria, and can be deployed without any changes to the target model.

\subsubsection{Experiments}

\begin{table*}[t]
\centering
\setlength{\tabcolsep}{2.5pt}
\renewcommand{\arraystretch}{1.2}

\resizebox{\textwidth}{!}{
\begin{tabular}{@{}l ccc >{\columncolor{gray!10}}ccc ccc >{\columncolor{gray!10}}ccc ccc >{\columncolor{gray!10}}ccc@{}}
\toprule
 & \multicolumn{9}{c}{\textbf{Math}} 
 & \multicolumn{3}{c}{\textbf{Coding}} 
 & \multicolumn{3}{c}{\textbf{Science}} 
 & \multicolumn{3}{c}{\textbf{Logic}} \\
\cmidrule(lr){2-10} \cmidrule(lr){11-13} \cmidrule(lr){14-16} \cmidrule(lr){17-19}

\textbf{Method}
 & \multicolumn{3}{c}{\textbf{GSM8K}}
 & \multicolumn{3}{c}{\textbf{MATH500}}
 & \multicolumn{3}{c}{\textbf{AIME}}
 & \multicolumn{3}{c}{\textbf{HUMANEVAL+}}
 & \multicolumn{3}{c}{\textbf{GPQA-D}}
 & \multicolumn{3}{c}{\textbf{ARC-Challenge}} \\

 & Acc$\uparrow$ & Tok$\downarrow$ & Lat$\downarrow$
 & Acc$\uparrow$ & Tok$\downarrow$ & Lat$\downarrow$
 & Acc$\uparrow$ & Tok$\downarrow$ & Lat$\downarrow$
 & Acc$\uparrow$ & Tok$\downarrow$ & Lat$\downarrow$
 & Acc$\uparrow$ & Tok$\downarrow$ & Lat$\downarrow$
 & Acc$\uparrow$ & Tok$\downarrow$ & Lat$\downarrow$ \\
\midrule

\multicolumn{19}{l}{\textit{\textbf{Qwen3-4B-Thinking-2507}}} \\
\textit{Think}
 & 95.3 & 1414 & 155.6 & 96.6 & 6719 & 530.1 & 86.7 & 19577 & 243.3 & 90.9 & 5079 & 175.3 & 68.7 & 9041 & 325.8 & 95.6 & 1812 & 156.5 \\

\textit{NoThink*}
 & 95.2 & 1631 & 204.2 & 96.6 & 6395 & 488.5 & 86.7 & 19816 & 243.2 & 88.4 & 4480 & 131.5 & 67.2 & 8833 & 276.8 & 95.1 & 1889 & 159.8 \\

\textit{DEER}
 & 94.3 & 960 & 230.3 & 94.4 & 4893 & 519.6 & 70.0 & 17838 & 218.6 & 86.6 & 4079 & 242.4 & 67.2 & 9053 & 505.2 & 94.6 & 1011 & 200.3 \\

\textit{EAGLE3}
 & 94.8 & 1408 & 140.3 & 96.6 & 6670 & 395.7 & 80.0 & 19792 & 187.3 & 87.2 & 5178 & 81.7 & 67.7 & 8975 & 212.2 & 95.7 & 1822 & 164.2 \\

\textbf{SpecExit}
 & \textbf{93.8} & \textbf{649} & \textbf{75.8}
 & \textbf{96.8} & \textbf{4777} & \textbf{367.9}
 & \textbf{90.0} & \textbf{17769} & \textbf{206.1}
 & \textbf{89.6} & \textbf{4319} & \textbf{58.4}
 & \textbf{68.7} & \textbf{7011} & \textbf{137.0}
 & \textbf{94.5} & \textbf{588} & \textbf{71.4} \\

\midrule
\multicolumn{19}{l}{\textit{\textbf{DeepSeek-R1-Distill-Llama-8B}}} \\

\textit{Vanilla}
 & 76.4 & 1008 & 629.4 & 81.8 & 6878 & 857.1 & 36.7 & 22170 & 307.0 & 74.4 & 6287 & 445.5 & 43.6 & 8857 & 574.0 & 49.9 & 1917 & 628.5 \\

\textit{NoThink}
 & 54.6 & 233 & 22.2 & 55.2 & 1643 & 262.8 & 10.0 & 8744 & 184.1 & 46.3 & 472 & 7.3 & 26.8 & 1200 & 166.6 & 12.6 & 135 & 13.6 \\

\textit{DEER}
 & 74.7 & 710 & 484.8 & 80.8 & 3533 & 973.3 & 40.0 & 15619 & 272.3 & 79.3 & 4206 & 269.2 & 40.9 & 8492 & 521.5 & 47.5 & 1029 & 531.3 \\

\textit{EAGLE3}
 & 79.3 & 976 & 276.9 & 80.8 & 6172 & 593.6 & 30.0 & 25686 & 228.1 & 78.7 & 5312 & 346.5 & 43.9 & 8749 & 420.1 & 59.2 & 1378 & 496.4 \\

\textbf{SpecExit}
 & \textbf{75.3} & \textbf{333} & \textbf{112.6}
 & \textbf{80.6} & \textbf{1968} & \textbf{348.3}
 & \textbf{36.7} & \textbf{8160} & \textbf{176.0}
 & \textbf{81.7} & \textbf{3105} & \textbf{118.1}
 & \textbf{46.0} & \textbf{6849} & \textbf{307.5}
 & \textbf{50.3} & \textbf{500} & \textbf{253.7} \\

\bottomrule
\end{tabular}
}
\caption{Performance comparison of various reasoning methods on mathematical, scientific, general, and coding benchmarks. ``Acc'' denotes accuracy, ``Tok'' denotes token count, and ``Lat'' denotes total end-to-end latency. $\uparrow$ indicates higher is better; $\downarrow$ indicates lower is better.}
\label{tab:specexit_main_results}
\end{table*}

We evaluate SpecExit on a diverse set of reasoning benchmarks spanning mathematics (GSM8K, MATH500, AIME), coding (HumanEval+), science (GPQA-D), and logical reasoning (ARC-Challenge). Experiments are conducted on two representative large reasoning models, Qwen3-4B-Thinking-2507 and DeepSeek-R1-Distill-Llama-8B, and compared against standard reasoning and inference-time acceleration baselines, including speculative decoding with EAGLE3. Table~\ref{tab:specexit_main_results} reports accuracy, generated token count, and end-to-end latency.

Across all benchmarks and models, SpecExit consistently achieves substantial reductions in reasoning length and inference latency while maintaining comparable accuracy. For Qwen3-4B-Thinking-2507, SpecExit reduces generated tokens by up to 54\% on GSM8K and 53\% on ARC-Challenge, translating into nearly 2$\times$ end-to-end latency reduction compared to the EAGLE3 baseline. On DeepSeek-R1-Distill-Llama-8B, the effect is even more pronounced: SpecExit shortens reasoning trajectories by up to 66\% and delivers up to 2.5$\times$ latency speedup on GSM8K, despite aggressive pruning of intermediate reasoning steps.

Importantly, these efficiency gains incur only marginal accuracy differences, confirming that SpecExit primarily removes redundant reasoning rather than truncating essential computation. In contrast to heuristic or rule-based early stopping methods, SpecExit achieves a more favorable trade-off between reasoning depth and latency by leveraging learned exit signals integrated into the speculative decoding pipeline. Overall, the results demonstrate that SpecExit is an effective and practical mechanism for accelerating large reasoning model inference without sacrificing task performance.

\section{Sparse Attention and Token Pruning}
In this section, we present our approach to optimize the computational efficiency of LLMs and MLLMs through structural and dynamic sparsity. As sequence lengths continue to scale, the self-attention mechanism becomes a primary bottleneck due to its $O(n^2)$ complexity. To address this challenge, AngelSlim proposes a two-pronged strategy: 

\textbf{Sparse Attention for Long-Sequence Prefill:} A framework specifically optimized to accelerate the KV cache generation stage in LLMs by exploiting the intrinsic sparsity of attention maps.

\textbf{Token Pruning for Multimodal Redundancy:} A universal framework tailored for VLMs and Speech LLMs that reduces the effective sequence length by eliminating redundant visual and audio tokens.

\begin{figure}[t]
    \centering
    \includegraphics[width=\linewidth]{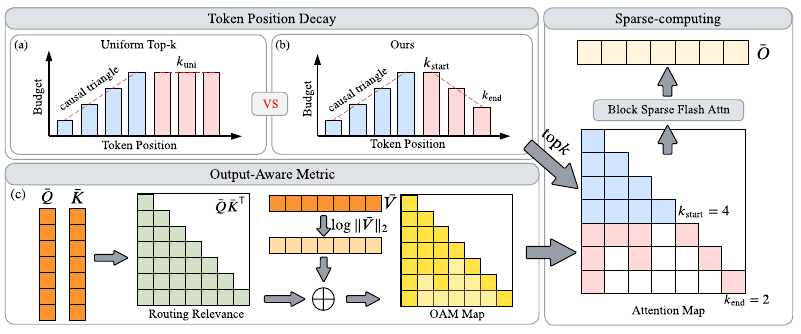}
    \caption{Architecture of the proposed Stem. 
    (a) Uniform Sparse Attention: Illustrates the traditional method applying a fixed top-k budget across all token positions, often leading to information loss in initial tokens.
    (b) Token Position-Decay (TPD): Contrasts with (a) by allocating dynamic budgets, assigning higher retention stability to initial tokens ("recursive anchors") to preserve causal structure.
    (c) Output-Aware Metric (OAM): A selection metric combining attention scores with value vector magnitudes to minimize output approximation error.}
    \label{fig:Stem}
\end{figure}
\subsection{Sparse Attention Framework}
To mitigate the quadratic computational complexity associated with long-context LLMs, AngelSlim provides a comprehensive sparse attention framework.
This framework is specifically optimized for the KV Cache Generation stage, enabling efficient prefill for sequences exceeding 100K tokens by exploiting the intrinsic sparsity within attention maps.

\subsubsection{Overall Framework}
\paragraph{A Versatile Library of Sparse Attention Patterns.} We provide a versatile sparse attention library to accommodate the diverse attention behaviors observed across mainstream LLMs.
The framework integrates several sophisticated strategies, primarily categorized into static and dynamic sparsity patterns.
Static sparsity utilizes structured patterns based on attention distribution heuristics to maintain a fixed sparsity mask during inference.
This primarily involves A-shape and Tri-shape configurations, and also supports Dilated and Strided attention to ensure a broad receptive field while significantly reducing the number of active tokens.
Dynamic sparsity involves identifying necessary tokens on-the-fly during inference, which first performs pattern computation to locate sparse regions and then executes sparse attention kernels accordingly.
Our framework supports a variety of advanced dynamic sparse algorithms, including MInference, xAttention, FlexPrefill, and Stem.
\paragraph{Decoupling Sparse Kernels from Model Architectures.} We implement a strict decoupling between various sparse kernels and model architectures to simplify the deployment of diverse sparse algorithms.
AngelSlim acts as a unified management layer that orchestrates various high-performance open-source kernels tailored for different sparse attention mechanisms.
Through a training-free and metadata-driven configuration system, researchers can flexibly apply optimal sparsity settings to specific layers or heads for models like Qwen3, Llama3.1, and the Hunyuan-Dense-Model series.
This design significantly accelerates the prefill stage, effectively reducing the time-to-first-token (TTFT) during long-context inference.
\paragraph{Automated Evaluation for Long-Context Capabilities.} We integrate an automated evaluation pipeline that systematically benchmarks sparsified models on mainstream long-context suites.
The system supports end-to-end evaluation on benchmarks such as LongBench for document understanding and RULER for testing retrieval robustness under high-ratio sparsity.
By providing detailed metrics on sparsity ratio, time-to-first-token (TTFT), and task accuracy, this pipeline enables users to identify the most efficient configuration for large-scale deployment.

\subsubsection{Algorithm}
\paragraph{Stem: Rethinking Causal Information Flow.} \begin{wraptable}{r}{0.48\columnwidth}
\vspace{-1em}
\centering
\renewcommand{\arraystretch}{1.08}
\setlength{\tabcolsep}{3.08pt}
\resizebox{\linewidth}{!}{
\begin{tabular}{@{}lcccccc|c@{}}
\toprule
Method & CC & FSL & MD1 & MD2 & SUM & SYN & AVG \\
\midrule
\multicolumn{8}{c}{\textbf{Qwen3-8B}} \\
\midrule
Dense & 19.09 & 63.10 & 11.23 & 15.06 & 20.63 & 62.92 & 32.01 \\
MINF  & 18.91 & 61.87 & 10.78 & 14.46 & 20.26 & 55.32 & 30.27 \\
FLEX  & 19.75 & 55.31 & 10.76 & 13.49 & \textbf{21.18} & 50.83 & 28.55 \\
XATTN & \textbf{21.03} & \textbf{62.30} & \textbf{11.42} & 14.64 & 20.47 & 52.92 & 30.46 \\
Stem  & 19.43 & 61.84 & 11.22 & \textbf{14.97} & 20.21 & \textbf{62.19} & \textbf{31.64} \\

\midrule
\multicolumn{8}{c}{\textbf{Llama-3.1-8B-Instruct}} \\
\midrule
Dense & 36.08 & 64.29 & 27.54 & 31.19 & 25.10 & 67.92 & 42.02 \\
MINF  & 35.43 & 62.98 & 25.42 & 29.00 & 25.12 & \textbf{68.41} & 41.06 \\
FLEX  & 34.65 & 59.51 & 17.17 & 21.62 & 24.46 & 59.10 & 36.09 \\
XATTN & \textbf{37.23} & \textbf{63.61} & 22.15 & 28.23 & \textbf{25.30} & 50.95 & 37.91 \\
Stem  & 35.86 & 62.89 & \textbf{26.33} & \textbf{30.53} & 24.93 & 68.32 & \textbf{41.48} \\

\midrule
\multicolumn{8}{c}{\textbf{DeepSeek-V3.2 (trained sparse)}} \\
\midrule
DSA & \textbf{32.42} & 41.85 & \textbf{51.83} & 48.30 & \textbf{22.28} & 60.35 & 42.84 \\
DSA + Stem & 32.02 & \textbf{44.03} & 51.67 & \textbf{48.44} & 21.97 & \textbf{60.85} & \textbf{43.16} \\

\bottomrule
\end{tabular}
}
\caption{LongBench accuracy of Stem.}
\vspace{-6pt}
\label{tab:longbench-merged-nobudget}
\end{wraptable}
We introduce Stem, a plug-and-play, training-free sparsity module that prunes attention in a way that better matches causal information flow during long-context prefill. %
As illustrated in Fig.~\ref{fig:Stem}(a), a common baseline is uniform sparse attention: applying the same top-$k$ budget to every query position, which often over-prunes the beginning of the sequence and discards information that many later tokens repeatedly depend on. %
Stem addresses this with two simple, report-friendly ideas shown in Fig.~\ref{fig:Stem}(b)(c). %
First, Token Position-Decay (TPD) (Fig.~\ref{fig:Stem}(b)) allocates a non-uniform budget across positions: early tokens are treated as ``recursive anchors'' and receive higher retention stability, while the budget gradually decays toward later tokens where redundancy is typically higher and aggressive sparsification is safer. %
Second, Output-Aware Metric (OAM) (Fig.~\ref{fig:Stem}(c)) changes how we decide what to keep: instead of selecting tokens purely by attention affinity, we also account for the effective contribution of the corresponding Value states, so tokens that look “high-score” but carry weak value signal are less likely to be selected, and tokens with meaningful value contribution are prioritized to reduce output distortion. %
Together, TPD provides a position-aware pruning schedule and OAM provides a contribution-aware selection rule, so Stem can preserve the critical early-sequence structure while still delivering large compute savings under high sparsity in long-context workloads. %

\begin{wrapfigure}{r}{0.48\columnwidth}
\vspace{-12pt}
\centering
\includegraphics[width=\linewidth]{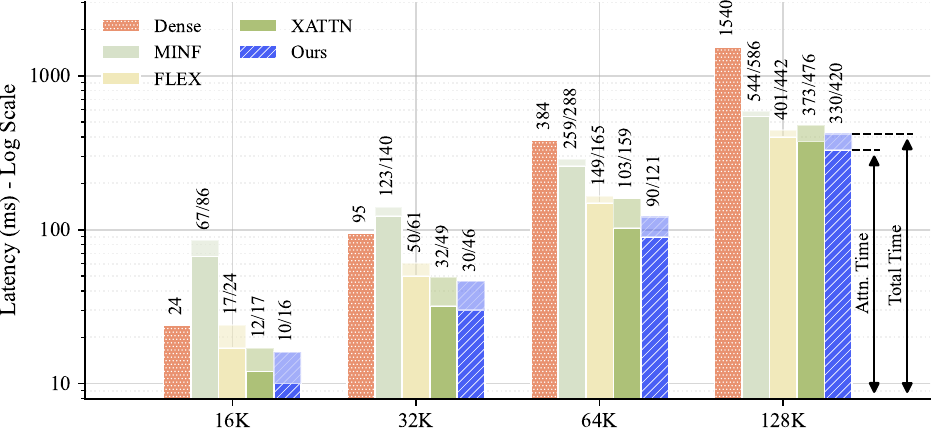}
\caption{Latency comparison (ms) on Nvidia GPU. Results are reported as Attention Kernel Time/Total Time.}
\label{fig:latency}
\vspace{-10pt}
\end{wrapfigure}

Effectiveness Across Dense and Trained-Sparse Backbones. Table~\ref{tab:longbench-merged-nobudget} reports LongBench accuracy across task families, including code completion (CC), few-shot learning (FSL), multi-document QA (MD1 and MD2), summarization (SUM), and synthetic tasks (SYN), with AVG as the overall average. 
In training free deployment on dense backbones, Stem consistently achieves stronger overall accuracy than prior sparse baselines on both Qwen3 and Llama, showing that it can be inserted into existing models without retraining and remain reliable. 
Stem also transfers to models with trained sparse attention. 
On DeepSeek-V3.2~\citep{liu2024deepseek}, which uses DSA as its native sparse attention, applying Stem on top of DSA maintains comparable performance, indicating that Stem can refine token selection and remove residual redundancy without changing the trained model. 
Figure~\ref{fig:latency} further shows that Stem consistently reduces long-context prefill latency, since its metric computation remains lightweight and the position-decay schedule enables faster sparse execution. 

\paragraph{Other Sparse Attention Strategies.} Beyond Stem, our framework provides comprehensive support for a wide range of common sparse attention Strategies, including FlexPrefill~\citep{flexprefill}, MInference~\citep{minference}, and XAttention~\citep{xattention}. %
By integrating these techniques into a unified codebase, we streamline the experimental workflow and ensure fair comparisons among different sparsity paradigms.

\subsection{Token Pruning}

In this subsection, we present our universal framework for token pruning and merging in multimodal settings, along with two novel algorithms tailored to different modalities: IDPruner for visual token pruning and Samp for audio token pruning and merging.

\subsubsection{Framework}

\begin{figure*}[t]
    \centering
    \includegraphics[width=\linewidth]{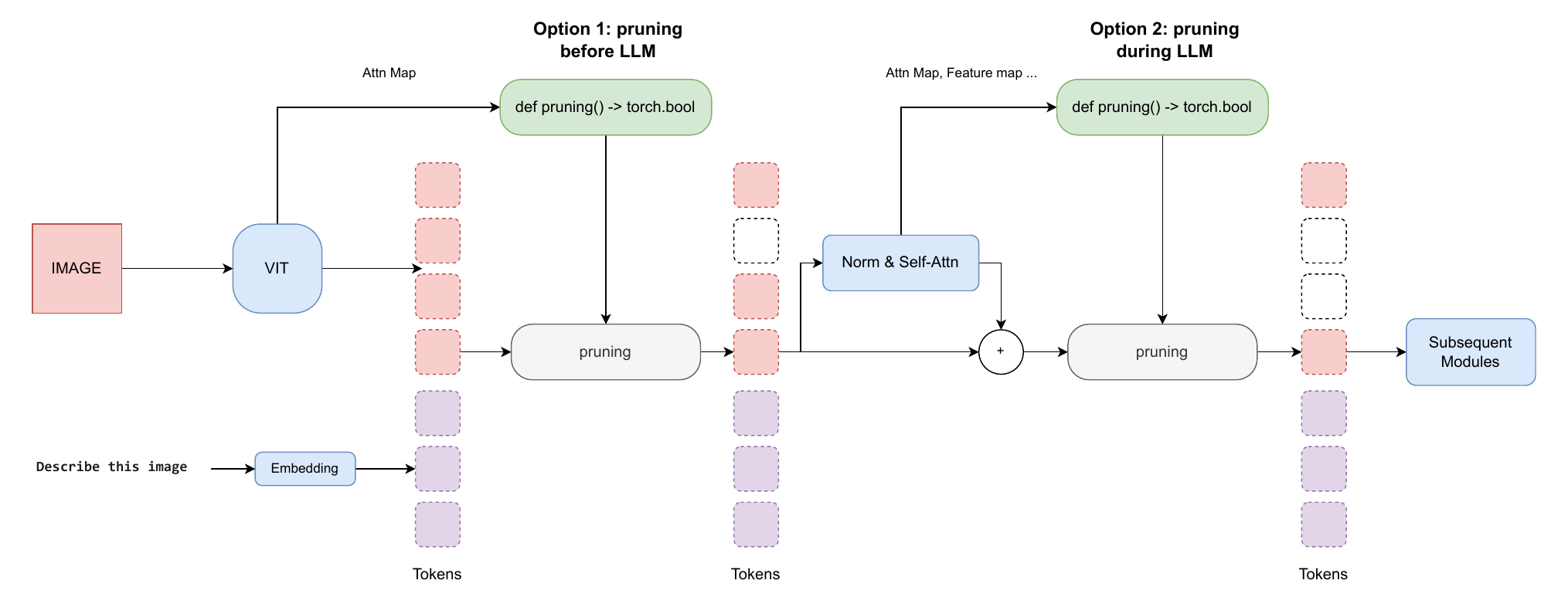}
    \caption{Architecture of the Universal Metadata-Driven Token Pruning Framework. The framework streamlines the pruning process by supporting two flexible compression schedules: Option 1 (Global Pruning) reduces the sequence length before entering the LLM by utilizing metadata from the Vision Tower (e.g., Attention Maps); Option 2 (Layer-wise Pruning) enables incremental sparsification between Transformer blocks based on intermediate states such as feature maps. The core design abstracts the pruning logic into a standardized interface (\texttt{def pruning() -> torch.bool}), effectively decoupling the algorithmic strategy from the underlying model architecture.}
    \label{fig:PruningFramework}
\end{figure*}

In this work, we present a universal metadata-driven framework designed for token pruning and merging in Multi-modal Large Language Models (MLLMs).
As illustrated in Fig.~\ref{fig:PruningFramework}, our architecture provides a standardized interface to simplify the overall token optimization process.
This approach allows researchers to evaluate whether model performance is effectively maintained across various benchmarks while simultaneously tracking metrics such as theoretical computational speedup and inference latency.

\paragraph{Decoupling Pruning Strategies from Model Architectures.} We implement a strict decoupling between pruning algorithms and specific model implementations to simplify the development cycle.
Algorithm researchers only need to define the core pruning strategy within a standalone function that outputs a boolean mask based on runtime context.
This design successfully isolates developers from the intricate details of the underlying model architectures.
Once the mask is generated, the framework automatically handles essential downstream operations, such as slicing hidden states and synchronizing metadata—including attention masks, position embeddings, and KV cache position vectors.
For strategies requiring intermediate information such as attention maps, the necessary data can be requested through YAML configurations.
The framework dynamically captures these tensors during the forward pass and passes them as arguments to the pruning function.

\paragraph{Integrated End-to-End Evaluation Pipeline.} We also provide a comprehensive solution that unifies algorithm development with systematic performance benchmarking.
By integrating mainstream multi-modal evaluation suites directly into the compression engine, we have created a fully automated end-to-end pipeline.
Through a unified interface, users can systematically evaluate optimized models across a suite of downstream tasks—such as TextVQA, MME, and DocVQA—at various pruning ratios.
The system automatically generates detailed reports that present both task accuracy and theoretical computational speedup.
This closed-loop design effectively reduces engineering overhead and accelerates the transition of advanced compression methods from theoretical research to practical deployment.

\subsubsection{Visual Token Pruning Algorithms}

\begin{figure}[t]
    \centering
    \includegraphics[width=\linewidth]{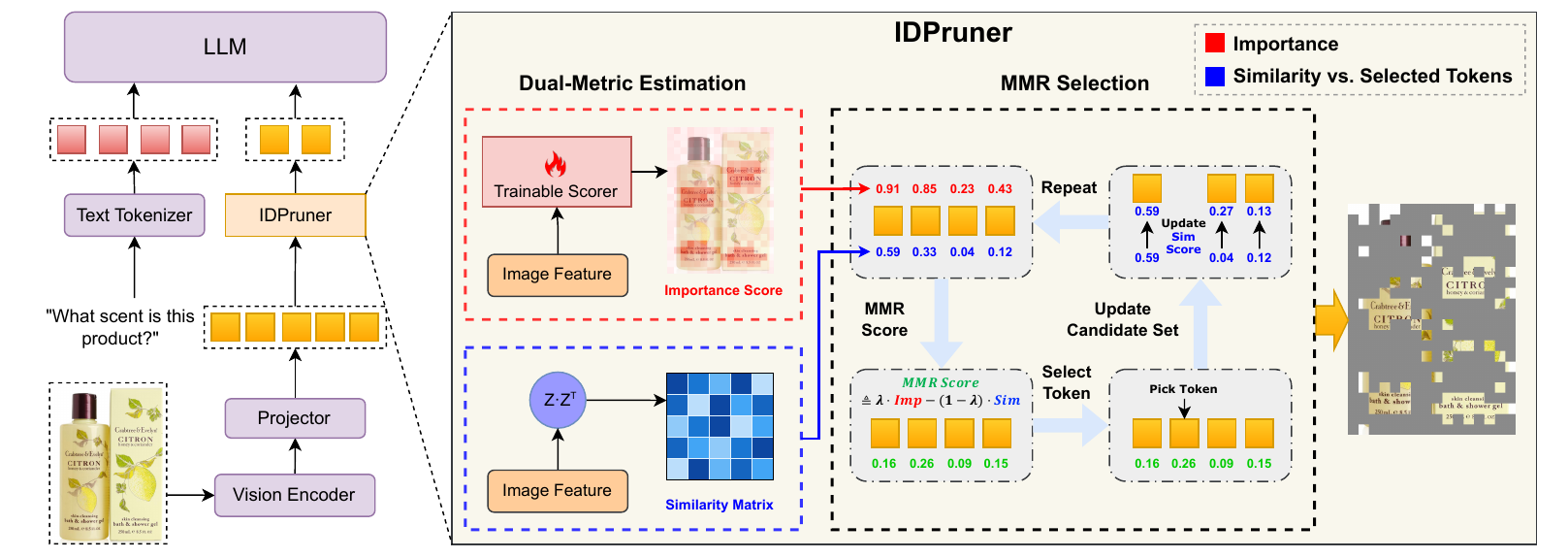}
    \caption{Architecture of the proposed IDPruner with importance-diversity optimization.
    By computing Importance Scores and a Similarity Matrix, IDPruner dynamically balances token saliency and semantic diversity through an MMR-based iterative selection process.
    }
    \label{fig:idpruner_overview}
\end{figure}
\begin{table*}[t]
\resizebox{\textwidth}{!}{
\begin{tabular}{lcccccccccccc}
\toprule
\multirow{2}{*}{Method} & AI2D & ChartQA & DocVQA & MMB\textsuperscript{CN} & MMB & MME & MMStar & OCRBench & POPE & SQA & VQA\textsuperscript{Text} & \multirow{2}{*}{Avg} \\
 & \textit{EM} & \textit{Relaxed} & \textit{Anls} & \textit{Score} & \textit{Score} & \textit{Score} & \textit{Avg} & \textit{Acc} & \textit{Acc} & \textit{EM} & \textit{EM} & \\
\midrule
Baseline & 82.48 & 83.68 & 94.90 & 80.41 & 83.08 & 1702 & 61.88 & 85.30 & 87.80 & 88.45 & 82.74 & 100.0\% \\
\midrule
\multicolumn{13}{c}{\cellcolor{gray!15} Retain 25\% Tokens (75\% Compression Ratio) } \\
FastV & 75.68 & 68.20 & 81.20 & 73.20 & 76.12 & 1636 & 51.08 & 43.00 & 85.20 & 83.49 & 80.06 & 87.16\% \\
VisionZip & 77.40 & 67.20 & 71.48 & 76.12 & 78.78 & 1637 & 54.86 & 46.50 & 85.76 & 83.99 & 76.21 & 87.55\% \\
HiPrune & 77.49 & 68.60 & 73.52 & 76.03 & 78.09 & 1619 & 54.43 & 47.10 & 86.02 & 84.18 & 76.43 & 87.80\% \\
VisionSelector & 79.60 & 72.00 & 93.24 & 75.86 & 78.78 & 1688 & 55.78 & 72.50 & 86.74 & 85.08 & 80.39 & 94.22\% \\
DivPrune & 77.98 & 62.00 & 85.32 & 75.77 & 77.84 & 1650 & 52.97 & 58.40 & 85.88 & 83.94 & 75.88 & 89.26\% \\
DART & 74.35 & 60.80 & 78.90 & 73.88 & 76.72 & 1625 & 52.90 & 46.00 & 84.34 & 84.33 & 71.68 & 85.74\% \\
VisPruner & 77.62 & 68.04 & 77.39 & 75.69 & 78.87 & 1657 & 54.01 & 48.70 & 85.68 & 84.18 & 75.17 & 88.31\% \\
SCOPE & 78.92 & 71.20 & 85.40 & 77.75 & 79.38 & 1684 & 56.86 & 61.70 & 86.78 & 85.23 & 79.66 & 92.51\% \\
\rowcolor{HYLightBlue} \textbf{IDPruner} & 80.51 & 74.32 & 93.16 & 76.63 & 79.73 & 1695 & 56.49 & 74.00 & 87.06 & 85.52 & 80.83 & 95.18\% \\
\midrule
\multicolumn{13}{c}{\cellcolor{gray!15} Retain 10\% Tokens (90\% Compression Ratio) } \\
FastV & 67.23 & 39.48 & 51.90 & 53.26 & 55.58 & 1332 & 38.02 & 24.10 & 76.31 & 79.28 & 72.59 & 68.07\% \\
VisionZip & 70.60 & 41.56 & 37.94 & 66.67 & 71.05 & 1462 & 45.19 & 23.40 & 81.06 & 83.24 & 61.06 & 71.84\% \\
HiPrune & 69.82 & 43.96 & 39.89 & 67.44 & 70.88 & 1438 & 45.04 & 23.70 & 80.70 & 82.65 & 62.51 & 72.22\% \\
VisionSelector & 74.81 & 62.68 & 87.00 & 68.99 & 71.65 & 1569 & 46.93 & 55.50 & 82.69 & 81.95 & 74.52 & 85.39\% \\
DivPrune & 70.11 & 41.36 & 66.20 & 69.42 & 72.16 & 1529 & 44.46 & 31.80 & 81.91 & 80.96 & 62.72 & 76.09\% \\
DART & 67.88 & 34.84 & 49.86 & 63.92 & 67.35 & 1451 & 42.93 & 24.30 & 79.70 & 80.96 & 54.06 & 69.80\% \\
VisPruner & 69.88 & 42.68 & 50.85 & 66.84 & 70.96 & 1442 & 44.14 & 24.40 & 81.03 & 81.11 & 59.66 & 72.60\% \\
SCOPE & 71.63 & 50.04 & 56.45 & 71.22 & 75.43 & 1608 & 48.74 & 34.10 & 84.10 & 82.25 & 70.61 & 79.35\% \\
\rowcolor{HYLightBlue} \textbf{IDPruner} & 75.16 & 62.48 & 85.98 & 71.65 & 74.66 & 1618 & 47.48 & 53.90 & 85.43 & 82.80 & 74.43 & 86.47\% \\
\bottomrule
\end{tabular}
}
\centering
\caption{Performance comparison of different pruning methods on Qwen-2.5-VL-7B-Instruct.}
\label{tab:results_qwen2.5_7b_vl}
\end{table*}

\paragraph{IDPruner: A State-of-the-Art Pruning Method.} We propose IDPruner, a novel strategy that reformulates visual token pruning as a re-ranking problem using the Maximal Marginal Relevance (MMR) algorithm.
Unlike previous methods that rely on single metrics, IDPruner explicitly models the interplay between token importance and semantic diversity to achieve a Pareto-optimal balance.
As illustrated in Figure~\ref{fig:idpruner_overview}, the algorithm iteratively selects tokens that maximize normalized saliency scores while minimizing redundancy relative to the already selected subset.
Extensive experiments across diverse benchmarks and model architectures demonstrate that IDPruner achieves state-of-the-art performance, as shown in Table \ref{tab:results_qwen2.5_7b_vl}.

\paragraph{Other Pruning Strategies.} Beyond IDPruner, our framework provides comprehensive support for a wide range of common pruning and merging methods, including VisionSelector \cite{zhu2025visionselectorendtoendlearnablevisual}, FastV \cite{chen2024image}, DivPrune \cite{Alvar2025DivPruneDV}, DART \cite{Wen2025StopLF}, VisionZip \cite{yang2024visionziplongerbetternecessary}, VisPruner \cite{VisPruner}, and SCOPE \cite{Deng2025SCOPESO}.
The unified implementation of these diverse methodologies enables researchers to perform systematic comparisons across different pruning paradigms with minimal effort.

\begin{figure}[t]
    \centering
    \includegraphics[width=0.9\linewidth]{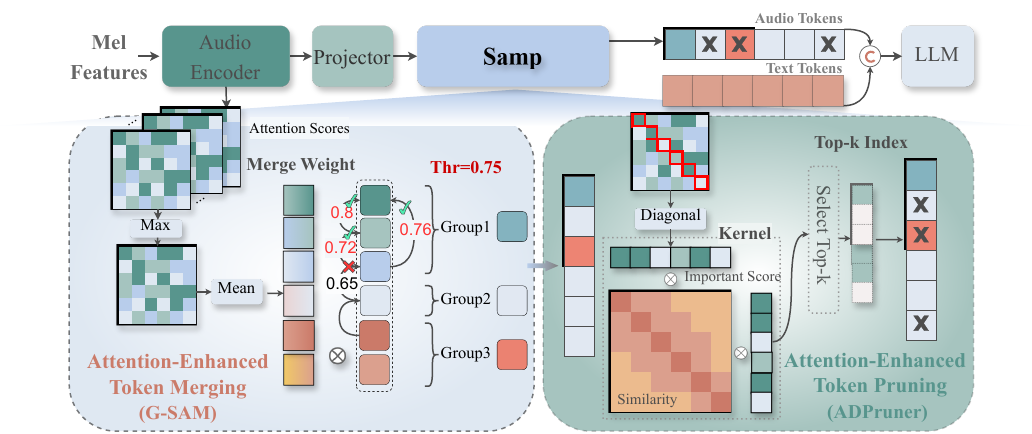}
    \caption{Architecture of the proposed \textbf{Samp} with adaptive merge–prune fusion. By introducing a similarity threshold, \textbf{Samp} dynamically balances token merging and pruning, according to speech temporal structures. Similar tokens are first adaptively merged to maximize information retention, followed by pruning to reduce redundancy and improve diversity under high compression rates.}
    \label{fig:Samp}
\end{figure}
\subsubsection{Audio Token Pruning Algorithms}

\paragraph{Samp: a plug-and-play, Similarity-Attention synergistically driven framework for joint token Merging and Pruning.}
Methods that integrate token pruning modules into LLM Transformer blocks often encounter compatibility bottlenecks, as highly optimized kernels like FlashAttention~\cite{dao2022flashattentionfastmemoryefficientexact} restrict access to internal attention scores. To maximize hardware efficiency and ensure compatibility with optimized inference kernels, we position the Samp module before the LLM. In contrast to approaches that require deep integration into Transformer blocks, our method relies solely on attention scores from a single audio encoder layer, which significantly reduces architectural coupling and memory overhead.
Tailored to the temporal dependencies and redundancy of speech tokens, we propose a two-part merging-pruning pipeline, as shown in Figure~\ref{fig:Samp}.

In the first merging stage, we group adjacent tokens based on intra-group global feature similarity and perform attention-guided weighted mergeing to collapse redundant segments while preserving critical semantic information.
Given the audio features  $\boldsymbol{H}_{a} \in \mathbb{R}^{N \times D}$ generated by the projection module preceding the LLM, we compute the pairwise cosine similarity across tokens to derive the similarity matrix $\boldsymbol{L}$. As illustrated in Fig. \ref{fig:Samp}, we fix the merge similarity threshold $\lambda$ and iterate through the audio token list $S=\{1, ..., N\}$ to sequentially incorporate multiple adjacent indices into a merged cluster. Let $s_i = \{k,...,k+m-1\}$ denote the set of audio tokens in the original merged cluster, where the cardinality of $s_i$  is $m$. We then compute the average similarity between the next adjacent token (with index $k+m$) and the tokens in $s_i$ , which is formally defined as:
\begin{equation}
l_{k+m,s_i}=\frac{1}{m}\sum_{t=k}^{k+m-1} \boldsymbol{L}_{k+m,t},\qquad
\boldsymbol{L}_{k+m,t} = \frac{\boldsymbol{H}_{a}^{k+m} \cdot \boldsymbol{\boldsymbol{H}}_{a}^{t}}{\|\boldsymbol{H}_{a}^{k+m} \| \cdot \|\boldsymbol{H}_{a}^{t} \|}.
\end{equation}
If $l_{k+m,s_i} \geq \lambda$,  we incorporate the token indexed by $k+m$ into $s_i$ ; otherwise, we initialize a new cluster $s_{i+1} =\{k+m\}$. The above procedure is repeated iteratively until all audio tokens are assigned to corresponding clusters.
Ultimately, we yield $K$ merged clusters, formally denoted as $S^*=\{s_i \vert  i=1,…,K\}$. To maximize the discriminative importance information encoded by attention, we compute the maximum value across the head dimension, and then derive the important score $W_j \in \mathbb{R} ^ N$ of each audio token. Then,  we aggregate the audio features of each subset $s_i$ in the cluster set $S^*$ (derived from the aforementioned clustering process) into a single token feature , ultimately yielding $K$ compressed audio tokens denoted as $ \tilde{\boldsymbol{H}}_a = \{\boldsymbol{\tilde{H}}_a^i \vert i = 1, ...,K \}$. and the corresponding index list set is updated to $\tilde{S}=\{1, ..., K\}$, in which:
\begin{equation}
\boldsymbol{\tilde{H}}_a^i = \frac{\sum_{}^{\vert s_i \vert} \boldsymbol{W}_j \cdot \boldsymbol{H}_a^j}{\sum_{}^{\vert s_i \vert} \boldsymbol{W}_j},\qquad
\boldsymbol{W}_{j} = \frac{1}{N}\sum_{n=0}^{N} \max\limits_{h} \boldsymbol{A}_{h,n,j}.
\label{eq:attn_merge}
\end{equation}
where $N$ denotes the length of the audio token sequence, $H$ stands for the number of attention heads, and attention scores $\boldsymbol{A} \in \mathbb{R} ^{H \times N \times N}$, $j \in s_i$, and $\boldsymbol{\tilde{H}}_a^i$ denotes the merged feature corresponding to the cluster subset $s_i$.
In the next pruning Stage, we implement a pruning kernel that integrates both attention and similarity information for diversity-driven pruning.
Specifically, we weight the original kernel matrix by the derived importance scores to construct a novel conditional kernel matrix:
\begin{equation}
    \boldsymbol {\hat{L}} = diag \left( \boldsymbol{ \hat A} \right) \cdot \boldsymbol{L} \cdot diag \left( \boldsymbol{ \hat A} \right),\qquad
    \boldsymbol{\hat{A}} = \frac{1}{H} \sum_{i=0}^{H} \boldsymbol{A}_{h,n,j}.
    \label{eq:dpp_kernel}
\end{equation}
We then obtain the optimal subset via MAP inference \cite{chen2024expanding}. By incorporating a similarity thresholding mechanism, \textbf{Samp} adaptively calibrates the ratio between merging and pruning for each individual sample, achieving a dynamic equilibrium between compression aggressiveness and task-specific accuracy.
\begin{table*}[t]
\centering
\normalsize  %
\setlength{\tabcolsep}{5pt}  %
\resizebox{\linewidth}{!}{%
\begin{tabular}{@{}l|ccccccccccc@{}}  %
\toprule  %
\multirow{2}{*}{\textbf{Method}} & \multicolumn{4}{c}{\textbf{LibriSpeech}} & \multicolumn{2}{c}{\textbf{Fleurs}} & \multirow{2}{*}{\textbf{AISHELL-1}} & \multirow{2}{*}{\textbf{AISHELL-2}} & \multicolumn{2}{c}{\textbf{WenetSpeech}} & \multirow{2}{*}{\textbf{Average}} \\
\cmidrule(lr){2-5} \cmidrule(lr){6-7} \cmidrule(lr){10-11}  %
& dev\_clean & dev\_other & test\_clean & test\_other & zh & en & & & test-meeting & test-net & \\
\midrule
\rowcolor{gray!20}\textbf{Qwen2-Audio} & 1.67 & 3.65 & 1.74 & 4.03 & 3.63 & 5.20 & 1.52 & 3.08 & 8.40 & 7.64 & 4.06 \\
\rowcolor{gray!10}\multicolumn{12}{c}{Retain 60\% Tokens (40\% Compression Ratio)} \\
VisionZip & 7.31 & 10.35 & 7.08 & 10.10 & 8.02 & 6.85 & 6.99 & 13.85 & 17.88 & 23.75 & 11.22 \\
VisPruner & 7.42 & 9.74 & 7.20 & 9.69 & 7.00 & 8.39 & 6.91 & 9.67 & 13.37 & 14.07 & 9.35 \\
CDPruner & 4.22 & 6.05 & 4.18 & 6.53 & 4.88 & 7.17 & 2.70 & 4.62 & 12.29 & 10.90 & 6.35 \\
A-ToMe & 4.12 & 6.81 & 4.20 & 6.98 & 8.00 & 8.13 & 4.18 & 5.56 & 14.05 & 14.15 & 7.62 \\
FastAdaSP & 4.91 & 7.26 & 4.95 & 7.51 & 5.47 & 7.31 & 3.28 & 4.69 & 11.51 & 12.30 & 6.92 \\
\rowcolor{HYLightBlue} \textbf{Samp} & \textbf{2.59} & \textbf{5.00} & \textbf{2.72} & \textbf{5.02} & \textbf{4.37} & \textbf{5.94} & \textbf{2.69} & \textbf{4.42} & \textbf{11.05} & \textbf{10.11} & \textbf{5.39} \\
\midrule
\rowcolor{gray!20}\textbf{Kimi-Audio} & 1.23 & 2.39 & 1.38 & 2.45 & 2.87 & 4.92 & 0.61 & 2.57 & 6.33 & 5.39 & 3.01 \\
\rowcolor{gray!10}\multicolumn{12}{c}{Retain 60\% Tokens (40\% Compression Ratio)} \\
VisionZip & 6.35 & 7.94 & 5.93 & 7.71 & 7.63 & 8.91 & 6.84 & 10.01 & 14.65 & 19.51 & 9.55 \\
VisPruner & 5.36 & 6.90 & 5.07 & 6.75 & 5.73 & 7.82 & 4.36 & 7.65 & 13.70 & 18.56 & 8.19 \\
CDPruner & 5.70 & 6.96 & 5.44 & 6.78 & 6.73 & 8.66 & 5.07 & 9.21 & 14.95 & 18.41 & 8.79 \\
A-ToMe & 4.13 & 6.04 & 3.92 & 5.87 & 6.26 & 7.57 & 3.58 & 6.88 & 12.97 & 14.99 & 7.22 \\
FastAdaSP & 4.49 & 6.15 & 4.28 & 6.02 & 4.27 & 7.64 & 2.22 & 5.67 & 12.62 & 14.75 & 6.81 \\
\rowcolor{HYLightBlue} \textbf{Samp} & \textbf{2.32} & \textbf{3.67} & \textbf{2.49} & \textbf{3.75} & \textbf{4.37} & \textbf{6.26} & \textbf{2.05} & \textbf{3.99} & \textbf{11.47} & \textbf{12.35} & \textbf{5.27} \\
\midrule
\rowcolor{gray!20}\textbf{GLM-ASR-Nano} & 2.14 & 4.05 & 2.18 & 4.53 & 3.44 & 4.11 & 2.47 & 3.48 & 8.43 & 6.65 & 4.15 \\
\rowcolor{gray!10}\multicolumn{12}{c}{Retain 70\% Tokens (30\% Compression Ratio)} \\
VisionZip & 9.18 & 12.75 & 8.27 & 13.27 & 8.17 & 8.08 & 13.9 & 20.55 & 19.36 & 20.45 & 13.40 \\
VisPruner & 9.22 & 12.33 & 8.34 & 12.13 & 6.95 & 8.33 & 11.68 & 14.57 & 18.11 & 16.56 & 11.82 \\
CDPruner & 5.38 & 7.75 & 5.28 & 7.76 & 4.99 & 5.93 & 4.88 & 5.94 & 14.21 & 11.70 & 7.38 \\
A-ToMe & 4.55 & 8.04 & 4.66 & 7.83 & 5.08 & 5.77 & 5.40 & 5.95 & 13.16 & 12.08 & 7.25 \\
FastAdaSP & 6.45 & 10.34 & 6.50 & 10.41 & 4.34 & 6.70 & 6.42 & 7.02 & 16.10 & 14.10 & 8.84 \\
\rowcolor{HYLightBlue} \textbf{Samp} & \textbf{3.41} & \textbf{5.53} & \textbf{3.43} & \textbf{5.55} & \textbf{3.76} & \textbf{4.83} & \textbf{3.53} & \textbf{4.52} & \textbf{12.07} & \textbf{9.60} & \textbf{5.62} \\
\bottomrule
\end{tabular}
}
\caption{Performance comparison of different pruning methods on ASR dense tasks (Qwen2-audio / Kimi-Audio / GLM-ASR)}
\label{tab:speech_performance}
\end{table*}

\paragraph{Other Pruning Strategies.} Beyond \textbf{Samp}, our framework also provides pure merging methods like A-ToMe~\cite{atome} and FastAdaSP~\cite{fastadasp}, pure pruning methods like VisPruner~\cite{VisPruner} and CDPruner~\cite{zhang2025attentionsimilaritymaximizingconditional}, and a hybrid approach which is VisionZip~\cite{yang2024visionziplongerbetternecessary}. By comparing these methods, the advantages of our proposed approach can be clearly demonstrated, for example, by its performance on the ASR task in Table~\ref{tab:speech_performance}.

\section{Conclusion and Future Works}

In this technical report, we introduced \textbf{\modelname{}}, a unified and comprehensive framework designed to surmount the ``Inference Wall'' by synergizing disparate model compression paradigms. By integrating weight-level quantization, structural sparsity, speculative decoding, and multimodal token pruning into a cohesive, hardware-aware manifold, \modelname{} provides a robust pipeline for scaling large foundation models. Our empirical evaluations demonstrate that \modelname{} achieves state-of-the-art performance across all optimization axes. Specifically, our ultra-low-bit quantization strategies, including the 2-bit QAT and ternary quantization methods, achieve accuracy parity with higher-precision formats while delivering a $4\times$ speedup on edge devices. Furthermore, the Eagle3 speculative decoding framework and our training-free sparse attention library significantly enhance inference throughput and long-context efficiency. By decoupling algorithmic strategies from low-level system implementations, \modelname{} empowers both researchers and practitioners to deploy LLMs and MLMs at a fraction of the original computational cost without compromising reasoning integrity. Ultimately, this toolkit serves as a critical infrastructure for the democratization of large-scale AI, bridging the gap between theoretical algorithmic innovation and practical industrial application.

\section{Contributions and Acknowledgments}
\modelname{} are the result of the collective efforts of all members of Tencent HY \modelname{} team. Below, we will list contributors.

\paragraph{Contributors (Ordered by the last name):} 

Rui Cen, QiangQiang Hu, Hong Huang, Hong Liu, Song Liu, Xin Luo, Lin Niu, Yifan Tan, Decheng Wu, Linchuan Xie, Rubing Yang, Guanghua Yu, Jianchen Zhu

\paragraph{Corresponding Author:} Guanghua Yu (lucayu@tencent.com)

\newpage

\bibliographystyle{citation}
\bibliography{citation}

\end{document}